\documentclass[]{interact}

\usepackage{epstopdf}

\usepackage{natbib}
\bibpunct[, ]{(}{)}{;}{a}{}{,}
\renewcommand\bibfont{\fontsize{10}{12}\selectfont}

\usepackage{caption}
\usepackage{subcaption}
\usepackage{mathtools}
\usepackage{hyperref}
\usepackage{todonotes}
\usepackage{graphicx}
\usepackage{graphbox}
\usepackage{booktabs}
\usepackage{amsfonts}       
\usepackage{xurl}
\usepackage[nolist]{acronym}
\usepackage{algorithm}
\usepackage[noend]{algpseudocode}
\usepackage{setspace}
\usepackage[nolists,tablesfirst]{endfloat}

\theoremstyle{plain}

\theoremstyle{definition}

\theoremstyle{remark}

\begin{document}

\articletype{ARTICLE}

\title{Learning Heuristics for Transit Network Design and Improvement with Deep Reinforcement Learning}

\author{
\name{Andrew Holliday\textsuperscript{a}\thanks{Email: andrew.holliday@mail.mcgill.ca} and Ahmed El-Geneidy\textsuperscript{b} \and Gregory Dudek\textsuperscript{a}}
\affil{\textsuperscript{a}McGill University Center for Intelligent Machines, 3480 University Street, McConnell Engineering Building, Room 410, Montreal, Quebec, Canada, H3A 0E9 \\
\textsuperscript{b}McGill University School of Urban Planning, 815 rue Sherbrooke West, Macdonald-Harrington Building, Room 400, Montreal, Quebec, Canada, H3A 0C2}
}


\maketitle

\begin{abstract}
Planning a network of public transit routes is a challenging optimization problem.  Metaheuristic algorithms search through the space of possible transit networks by applying heuristics that randomly alter routes in a network.  Existing algorithms almost exclusively use heuristics that modify the network in purely random ways.  In this work, we explore whether we can obtain better transit networks using more intelligent heuristics, that modify networks according to a learned preference function instead of at random.  We use reinforcement learning to train graph neural nets to act as heuristics.  These neural heuristics yield improved results on benchmark synthetic cities with 70 nodes or more, and achieve new state-of-the-art results on the challenging Mumford benchmark.  They also improve upon a simulation of the real transit network in the city of Laval, Canada, achieving cost savings of up to 19\% over the city's existing transit network.
\end{abstract}

\begin{keywords}
Public transit; urban planning; neural networks; reinforcement learning; optimization
\end{keywords}


\section{Introduction}



The COVID-19 pandemic caused declines in transit ridership in cities around the world~\citep{liu2020impacts}, putting many transit agencies under pressure to reduce operating costs~\citep{kar2022public}.  By improving a transit network's spatial layout, it may be possible to reduce costs while improving service quality.  But network redesigns can be very costly and are disruptive for riders, so it is vital that agencies get a design (or re-design) right the first time.  Good algorithms for the \ac{NDP} can therefore be very useful to transit agencies.

The \ac{NDP} is the problem of designing a high-quality set of transit routes for a city. It is an NP-hard problem, and real-world cities typically have hundreds or even thousands of transit stops, making analytical optimization approaches infeasible.  The most successful approaches to the \ac{NDP} to-date have been metaheuristic algorithms.  These repeatedly apply heuristics that randomly change a network, and use a metaheuristic rule such as natural selection or metallic annealing to decide which changes to keep.  

Different heuristics used in these algorithms make different kinds of changes to networks, such as randomly removing a stop from a route, or randomly exchanging two stops on a route.  What most heuristics have in common is that they make changes purely at random, ignoring the particulars of the network or city.  In this work, we use \ac{DRL} to train neural-net-based heuristics to choose changes to a transit network based on information about the city and the current network.  We integrate these with a simple evolutionary algorithm to form what we call a Neural Evolutionary Algorithm, and we show that in challenging scenarios, this technique outperforms existing non-neural algorithms.

This paper expands on our prior work~\citep{holliday2023augmenting, holliday2024autonomous} in a number of ways.  We present new results obtained with an improved evolutionary algorithm and an improved \ac{GNN} model trained via \acl{PPO} instead of REINFORCE.  We compare these results with state-of-the-art methods on the widely-used Mandl~\citep{mandl1980evaluation} and Mumford~\citep{mumford2013new} benchmark cities, and find that on the most challenging benchmarks our method sets a new state of the art, improving on other methods by up to 4.8\%.

We also report on ablation studies that show the importance of different features of our method.  As part of these studies, we consider a novel non-neural heuristic, where shortest paths with common end-points are joined uniformly at random instead of according to the \ac{GNN}'s output.  Interestingly, we find that in the narrow case where we are only concerned with minimizing passenger travel time, this heuristic outperforms both the baseline heuristics of~\cite{nikolic2013transit} and our \ac{GNN} heuristic; in most other cases, however, our \ac{GNN} heuristic performs better.

Finally, we perform a case study, applying our neural heuristics to a simulation of the real-world city of Laval, Canada, showing that they can plan transit networks that improve on the city's existing transit system by a wide margin by three distinct optimization criteria.  These results show that neural heuristics may allow transit agencies to offer better service at less cost.

\section{Background and related work}\label{sec:related_work}

\subsection{Deep Learning for Optimization Problems}\label{subsec:rw_dlforopt}

Deep reinforcement learning refers to the use of `deep' artificial neural nets - that is, neural nets with many hidden layers - as \ac{RL} models.  \ac{RL} is a branch of machine learning concerned with learning from feedback that comes in the form of a numerical `reward' that indicates how good or bad the model's outputs are, so that its outputs change to maximize the reward it receives.  As documented in ~\cite{bengio2021machine}, there is growing interest in the application of deep \ac{RL} to \ac{CO} problems.

\cite{vinyals2015pointer} proposed a deep neural net model called a Pointer Network, and trained it via supervised learning to solve instances of the \ac{TSP}.  This work has been built on by~\cite{dai2017learningCombinatorial}, \cite{Kool2019AttentionLT}, \cite{sykora2020multi} and others, training similar neural net models with \acl{RL} to construct \ac{CO} solutions.  These have attained impressive performance on the \ac{TSP}, the \ac{VRP}, and other \ac{CO} problems.  More recently, \cite{mundhenk2021symbolic} train a \ac{RNN} via \ac{RL} to construct a starting population of solutions to a genetic algorithm, the outputs of which are used to further train the \ac{RNN}.  \cite{fu2021generalize} train a model on small \ac{TSP} instances and present an algorithm that applies the model to much larger instances.  \cite{choo2022simulation} present a hybrid algorithm of Monte Carlo Tree Search and Beam Search that draws better sample solutions for the \ac{TSP} and \ac{CVRP} from a neural net policy like that of \cite{Kool2019AttentionLT}.

These approaches all belong to the family of `construction' methods, which solve a \ac{CO} problem by starting with an `empty' solution and adding to it until it is complete - for example, in the \ac{TSP}, this would mean constructing a path one node at a time, starting with an empty path and stopping once the path includes all nodes.  The solutions from these neural construction methods come close to the quality of those from specialized algorithms such as Concorde~\citep{concordeTspSolver}, while requiring much less run-time to compute~\citep{Kool2019AttentionLT}.

By contrast with construction methods, `improvement' methods start with a complete solution and repeatedly modify it, searching through the solution space for improvements.  In the \ac{TSP} example, this might involve starting with a complete path, and swapping pairs of nodes in the path at each step to see if the path is shortened.  Because this search process can continue indefinitely, improvement methods are generally more computationally costly than construction methods, but can yield better results.  Evolutionary algorithms belong to this category.  

Some work~\citep{hottung2019neural, chen2019learning, d2020learning, wu2021learning, ma2021learning}, has considered training neural nets to choose the search moves to be made at each step of an improvement method. meanwhile \cite{kim2021learning} train one neural net to construct a set of initial solutions, and another to modify and improve them.  And more recently, \cite{ye2024deepaco} train a neural net to provide a heuristic score for choices in \ac{CO} problems in the context of an \acl{ACO} algorithm.  This work has shown impressive performance on the \ac{TSP}, \ac{VRP}, and similar \ac{CO} problems.

In most of the above work, the neural net models used are \aclp{GNN}, a type of neural net model that is designed to operate on graph-structured data \citep{bruna2013spectral,kipf2016semi,defferrard2016spectral,duvenaud2015convolutional}.  These have been applied in many other domains, including the analysis of large web graphs~\citep{ying2018webscale}, the design of printed circuit boards~\citep{mirhoseini2021graph}, and the prediction of chemical properties of molecules~\citep{duvenaud2015convolutional, gilmer2017quantum}. 
An overview of \acp{GNN} is provided by~\cite{battaglia2018relational}.  Like many \ac{CO} problems, the \ac{NDP} lends itself to being described as a graph problem, so we use \ac{GNN} models here as well.

\ac{CO} problems also have in common that it is difficult to find a globally optimal solution but easier to gauge the quality of a given solution.  As ~\cite{bengio2021machine} note, this makes \acl{RL} a natural fit to \ac{CO} problems.  Most of the work cited in this section uses \ac{RL} methods to train neural net models.

\subsection{Optimization of Public Transit}\label{subsec:rw_ptopt}

The above work all concerns a small set of classical \ac{CO} problems including the \ac{TSP} and \ac{VRP}.  Like the \ac{NDP}, some of these are NP-hard, and all can be described as graph problems.  But the \ac{NDP} resembles these classical problems much less than they do each other: transit routes can interact by transferring passengers, and the space of solutions for a problem instance of given size is much vaster in the \ac{NDP}.  For example, take the Mumford0 benchmark city described in \autoref{tab:dataset}.  With 30 nodes there are $30! \approx 2.7 \times 10^{32}$ possible \ac{TSP} tours, but for 12 routes with at least 2 and at most 15 stops each, there are approximately $2.3 \times 10^{11}$ possible routes and so ${2.3 \times 10^{11} \choose 12} \approx 4.3 \times 10^{127}$ possible transit networks, a factor of $10^{95}$ more.

The difference is further shown by the fact that the state-of-the-art on classic problems like the \ac{TSP} uses analytical and mathematical programming methods (such as \cite{concordeTspSolver}), but on the \ac{NDP}, such methods are not used.  These differences necessitate a special treatment, so our work deals only with the \ac{NDP}, as does most other recent work on the problem \citep{mumford2013new, john2014routing, kilic2014demand, islam2019heuristic, ahmed2019hyperheuristic, lin2022analysis, husselmann2023improved, zervas2024solving}.

While analytical and mathematical programming methods have been successful on small instances~\citep{vannes2003AnalyticRouteAndSchedule, guan2006AnalyticRoutePlanning}, they struggle to realistically represent the problem~\citep{guihaire2008transitReview, kepaptsoglou2009transitReview}.  Metaheuristic approaches, as defined by~\cite{sorensen2018history}, have thus been more widely applied, both for the \ac{NDP} and the related \acl{FSP} \citep{aksoy2024comparing}.

The most widely-used metaheuristics for the \ac{NDP} have been genetic algorithms, simulated annealing, and ant-colony optimization, along with hybrids of these methods~\citep{guihaire2008transitReview, kepaptsoglou2009transitReview, duran2022survey, yang2020application, husselmann2023improved}.  Recent work has also shown other metaheuristics can be used with success, such as sequence-based selection hyper-heuristics~\citep{ahmed2019hyperheuristic}, beam search~\citep{islam2019heuristic}, and particle swarms~\citep{lin2022analysis}.  Many different heuristics have been applied within these metaheuristic algorithms, but most have in common that they select among possible neighbourhood moves uniformly at random.  

An exception is \cite{husselmann2023improved}.  For two heuristics, the authors design a simple model of how each change the heuristic could make would affect the network's quality.  The heuristics then weight the probability of making different changes according to this model, with higher-quality changes being more likely.  The resulting heuristics obtain state-of-the-art results.  However, their simple model ignores passenger trips involving transfers and the impact of the user's preferences over different parts of the cost function.  By contrast, the method we propose {\bf learns} a model of changes' impacts to assign probabilities to those changes, and does so based on a richer set of inputs.

While neural nets have been used for predictive problems in urban mobility ~\citep{xiong1992transportation, rodrigue1997NNsForLandUseAndTransport, chien2002dynamic, jeong2004bus, akgungor2009NNsForAccidentPrediction, li2020graph, Wu31122025} and for other transit optimization problems such as scheduling, passenger flow control, and traffic signal control~\citep{zou2006lightrail, ai2022deep, Yan2023DistributedMD, jiang2018passengerInflow, Zhang31122023, wang2024large, Wu31122025}, they have not often been applied to the \ac{NDP}.  The same is true of \ac{RL}: \cite{Li31122023} review the work on the application of \ac{RL} to public transit, and find only one instance in which it was used for network design: the work of \cite{wei2020city}.  In this work, the authors train a neural net policy via \ac{RL} to design a single new metro route for the city of Xi'an, China.  They obtain good results in that case, but their model of the problem differs from ours in not requiring the network to connect all stations, and their method does not go beyond planning a single route. 

Two other recent examples are~\cite{darwish2020optimising} and \cite{yoo2023reinforcement}.  Both use \ac{RL} to design routes and a schedule for the Mandl benchmark~\citep{mandl1980evaluation}, a very small city, and both obtain good results.  \cite{darwish2020optimising} use a \ac{GNN} approach inspired by~\cite{Kool2019AttentionLT}; in our own work we experimented with a nearly identical approach to~\cite{darwish2020optimising}, but we found it did not scale beyond very small instances.  Meanwhile, \cite{yoo2023reinforcement} use tabular \ac{RL}, an approach that is practical only for small problem sizes.  

All three of these approaches also require a new model to be trained on each problem instance.  Our approach, by contrast, finds good solutions for \ac{NDP} instances of more than 600 nodes, and can be applied to instances unseen during model training.

\section{The Transit Network Design Problem}\label{sec:tndp}

\begin{figure}
    \centering
    \includegraphics[scale=0.40]{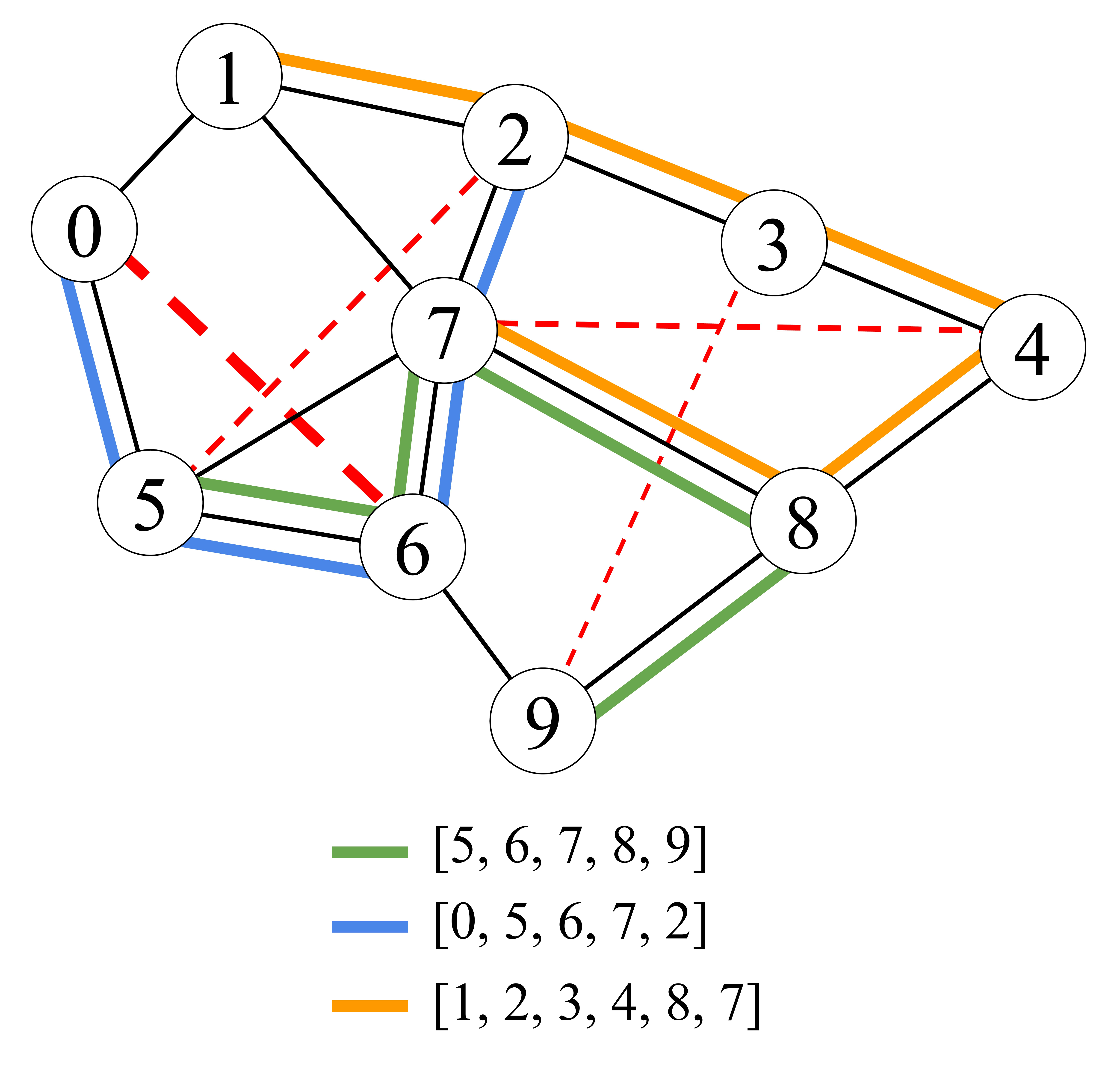}
    \caption{An example city graph with ten numbered nodes and three routes.  Link edges (representing streets or railways) are black, routes are in colour, and demands are shown by dashed red lines.  The edges of the three routes form a subgraph of the link graph $(\mathcal{N}, \mathcal{E}_s)$.  All node-pairs are connected by this subgraph, so the three routes form a valid transit network.  The demand between nodes 2 and 5 and between 0 and 6 can be satisfied directly by riding on the blue line, and the demand from 7 to 4 by the orange line.  The demand from 3 to 9 requires passengers to ride the orange line from node 3 to 8, and transfer to the green line to go from 8 to 9.}
    \label{fig:tndp}
\end{figure}

In the \acl{NDP}, a city is represented as an augmented graph:
\begin{equation}
 \mathcal{G} = (\mathcal{N}, \mathcal{E}_s, D)   
\end{equation}
This consists of a set $\mathcal{N}$ of $n$ nodes, representing candidate stop locations; a set $\mathcal{E}_s$ of link edges $(i, j, \tau_{ij})$ representing streets or railways that connect the nodes, with weights $\tau_{ij}$ indicating travel times on those links; and an $n \times n$ matrix $D$ giving the travel demand, in number of trips, between every pair of nodes in $\mathcal{N}$.  Our goal is to find a transit network $\mathcal{R}$ that minimizes a cost function $C: \mathcal{G}, \mathcal{R} \rightarrow \mathbb{R}^+$.  $\mathcal{R}$ is a set of routes where each route $r \in \mathcal{R}$ is a sequence of nodes in $\mathcal{N}$.  $\mathcal{R}$ is subject to the following constraints:
\begin{enumerate}
    \item $\mathcal{R}$ must satisfy all demand, providing some path over transit between every pair of nodes $(i,j) \in \mathcal{N}$ for which $D_{ij} > 0$.
    \item $\mathcal{R}$ must contain exactly $S$ routes ($|\mathcal{R}| = S$), where $S$ is given at the start.
    \item All routes $r \in \mathcal{R}$ must obey $MIN \leq |r| \leq MAX$, where $MIN$ and $MAX$ are parameters set by the user.
    \item Routes $r \in \mathcal{R}$ may not contain cycles; each node $i \in \mathcal{N}$ can appear in $r$ at most once.
\end{enumerate}
An example city graph with a transit network is shown in \autoref{fig:tndp}.

We deal here with the symmetric \ac{NDP}, meaning that all demand, links, and routes are the same in both directions:
\begin{align}
D &= D^\top \\
(i, j, \tau_{ij}) \in \mathcal{E}_s \; & \text{iff.} \; (j, i, \tau_{ij}) \in \mathcal{E}_s    
\end{align}
and all routes are traversed both forwards and backwards by vehicles on them.  This allows us to evaluate our method against the large number of other methods that evaluate on symmetric benchmark cities, but we note that extending our method to the asymmetric \ac{NDP} would be straightforward.

\subsection{Cost Functions}

Work on the \ac{NDP} usually considers two separate cost functions: the passenger cost $C_p$, and the operating cost $C_o$ \citep{nikolic2013transit, john2014routing, kilic2014demand, ahmed2019hyperheuristic, husselmann2023improved}.  We present the standard definitions of these terms used in this work below.

$C_p$ is defined as the average time of passenger trips over the network:
\begin{equation}
C_p(\mathcal{G}, \mathcal{R}) = \frac{\sum_{i,j} D_{ij}\tau_{\mathcal{R}ij}}{\sum_{i,j} D_{ij}}
\end{equation}
where $\tau_{\mathcal{R}ij}$ is the time of the shortest transit trip from $i$ to $j$ given $\mathcal{R}$, including a time penalty for each transfer between routes.

The operating cost is the total driving time of the routes:
\begin{equation}
C_o(\mathcal{G}, \mathcal{R}) = \sum_{r \in \mathcal{R}} \tau_r
\end{equation}
Where $\tau_r$ is the time needed to completely traverse a route $r$ in one direction.

We use these standard definitions in evaluating our algorithm.  However, our metaheuristic detailed in \autoref{subsec:evo_alg} requires a single cost function to operate, and our \ac{DRL} technique requires that this function not range much beyond $[-1, 1]$, as numerically large values can cause problems with neural net training.  So we define a unified cost function $C$ that combines passenger and operating costs and a third term that penalizes constraint violations that may occur during training.

In $C$, we use a modified passenger cost $C_p'$:

\begin{equation}
C_p'(\mathcal{G}, \mathcal{R}) = \frac{\sum_{i,j} D_{ij}(\delta_{\mathcal{R}ij}\tau_{\mathcal{R}ij} + (1-\delta_{\mathcal{R}ij})2\max\limits_{k,l} T_{kl})}{\sum_{i,j} D_{ij}}
\end{equation}
$\delta_{\mathcal{R}ij}$ is a delta function with value 1 if $\mathcal{R}$ provides a path from $i$ to $j$, and 0 if not.  $T$ is an $n \times n$ matrix of shortest-path travel times between every node pair.  Each demand that is not satisfied induces a penalty of $2\max\limits_{k,l} T_{kl}$, penalizing it, where $\max\limits_{k,l} T_{kl}$ is the maximum travel time between any two nodes in the graph.  If $\mathcal{R}$ satisfies constraint 1, then $\delta_{\mathcal{R}ij} = 1 \; \forall \; i,j$ and $C_p'$ reduces to the original $C_p$.

The constraint-violation cost $C_c$ is:
\begin{align}
C_c(\mathcal{G}, \mathcal{R}) =  F_{un} + F_s + 0.1 \delta_v
\end{align}

$F_{un}$ is the fraction of node pairs $(i,j)$ with $D_{ij} > 0$ for which $\mathcal{R}$ provides no path. $F_s$ is a smoothly increasing function of the extent to which the routes in $\mathcal{R}$ violate constraint 3:
\begin{align}
F_s = \frac{\sum_{r \in \mathcal{R}} \max(0, MIN - |r|, |r| - MAX)}{S * MAX}
\end{align}
$\delta_v$ is a delta function that has value 1 if $F_{un} > 0$ or $F_s > 0$, and has value 0 otherwise.  

This definition implies that when there are no constraint violations, $C_c = 0$.  We use fractional measures $F_{un}$ and $F_s$ so that usually, $C_c < 2$, as long as $MIN$ and $MAX$ scale roughly with $n$.  However, it has the drawback that if $n$ is large, and if $\mathcal{R}$ violates only a few constraints, $F_{un}$ and $F_s$ may become vanishingly small compared to $C_o$ and $C_p$.  This could lead to some constraint violations being permitted.  To prevent this, we include the term $0.1 \delta_v$, which ensures that even one violation will significantly increase overall cost $C(\mathcal{G}, \mathcal{R})$, no matter the size of the city graph.

$C_c$ does not penalize violations of constraints 2 and 4, because the \ac{MDP} detailed in \autoref{subsec:mdp} and the metaheuristic detailed in \autoref{subsec:evo_alg} are, by design, incapable of producing networks that violate these constraints.

The unified cost function is:
\begin{align}
C(\mathcal{G}, \mathcal{R}) = \alpha w_p C_p' + (1 - \alpha) w_o C_o + \beta C_c
\end{align}

The weight $\alpha \in [0, 1]$ controls the trade-off between passenger and operating costs, while $\beta$ is the constraint violation weight.  $w_p$ and $w_o$ are re-scaling constants chosen so that $w_p C_p'$ and $w_o C_o$ both vary roughly over the range $[0, 1)$ for different $\mathcal{G}$ and $\mathcal{R}$.  The values used are:
\begin{align}
    w_p &= (\max_{i,j}T_{ij})^{-1} \\
    w_o &= (S\max_{i,j}T_{ij})^{-1}
\end{align}

\subsection{Markov Decision Process Formulation}\label{subsec:mdp}

A \acf{MDP} is a formalism used to define problems in \ac{RL}~\cite[Chapter~3]{sutton2018reinforcement}.  In an \ac{MDP}, an {\bf agent} interacts with an {\bf environment} over a series of {\bf timesteps} $t$.  At each timestep $t$, the environment is in a \textbf{state} $s_t$, and the agent observes the state and takes some \textbf{action} $a_t \in \mathcal{A}_t$, where $\mathcal{A}_t$ is the set of available actions at that timestep.  The environment then transitions to a new state $s_{t+1}$ according to the state transition distribution $P(s_{t+1} | s_t, a_t)$, and the agent receives a \textbf{reward} $R_t \in \mathbb{R}$ according to the reward function $R_t = f_R(s_t, a_t, s_{t+1})$.  The agent chooses actions according to its \textbf{policy} $\pi(a_t|s_t)$, which is a probability distribution over $\mathcal{A}_t$ given $s_t$.  In \ac{RL}, the goal is to find a policy $\pi$ that maximizes the return $G_t$, defined as the time-discounted sum of rewards: 
\begin{equation}\label{eqn:return}
G_t = \sum^{t_\text{end}}_{t'=t} \gamma^{t' - t} R_{t'}
\end{equation}

Where $\gamma \in [0,1]$ is a parameter that discounts rewards farther in the future, and $t_\text{end}$ is the final time-step of the \ac{MDP}.  The sequence of states visited, actions taken, and rewards received from $t=1$ to $t_\text{end}$ constitutes one \textbf{episode} of the \ac{MDP}.

\begin{figure}
	\centering
	\includegraphics[width=\columnwidth]{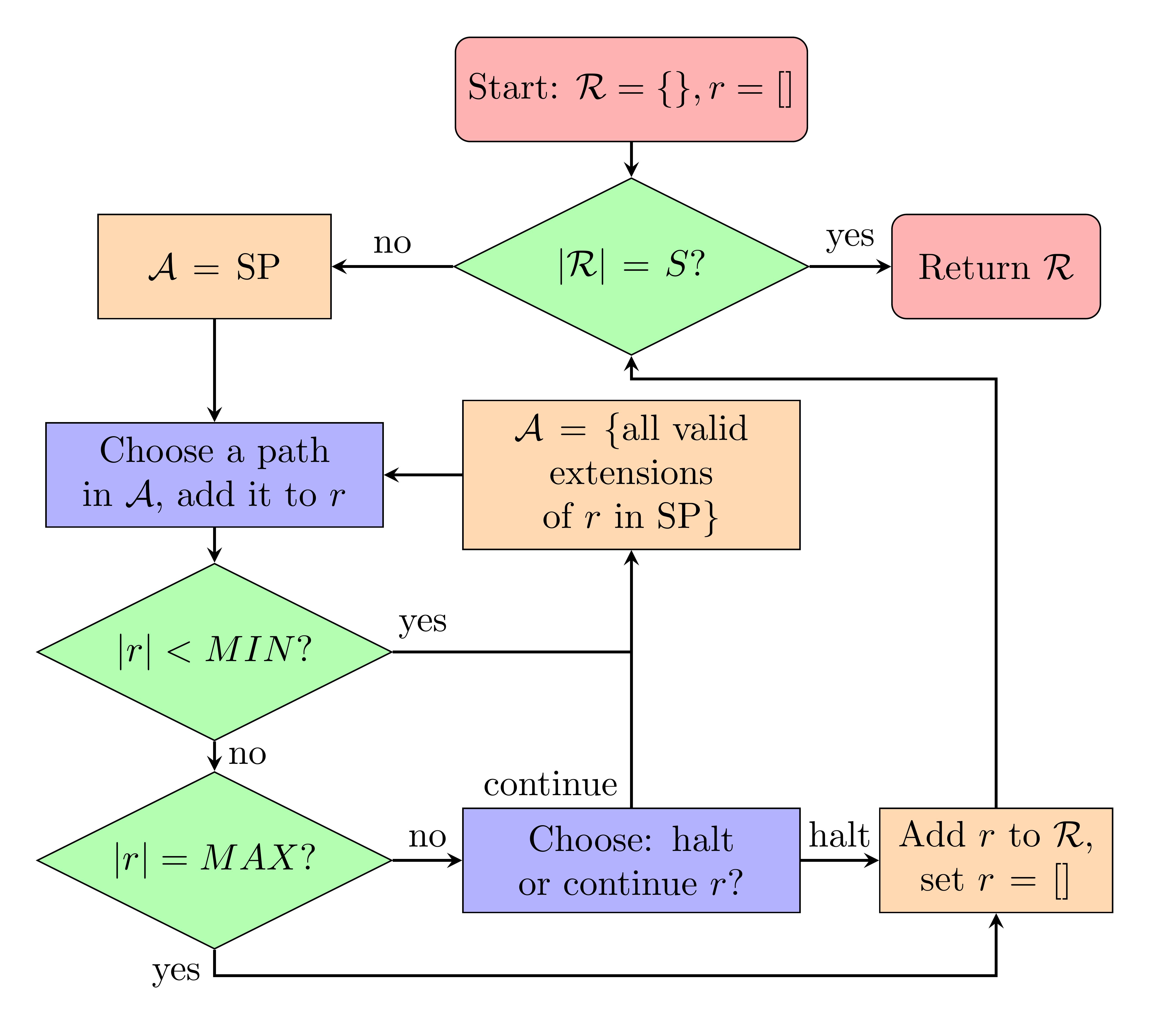}
	\caption{A flowchart of the transit network construction process defined by our \ac{MDP}.  Blue boxes indicate points where the timestep $t$ is incremented and the agent selects an action.  Red nodes are the beginning and ending of the process.  Green nodes are hard-coded decision points.  Orange nodes show updates to the state $s$ and action space $\mathcal{A}$.}
	\label{fig:mdp_flowchart}
\end{figure}

We here define an \ac{MDP} that constructs a transit network for the \acl{NDP} (\autoref{fig:mdp_flowchart}).  As shown in \autoref{eqn:state}, the state $s_t$ is composed of the set of finished routes $\mathcal{R}_t$, and an in-progress route $r_t$ which is currently being planned:
\begin{align}\label{eqn:state}
    s_t = (\mathcal{R}_t, r_t)
\end{align}
The starting state $s_1$ is $(\mathcal{R}_1 = \{\}, r_1 = [])$.  The \ac{MDP} alternates at every timestep between two modes: on odd-numbered timesteps, the agent selects an extension to the route $r_t$ that it is currently planning; on even-numbered timesteps, the agent chooses whether or not to stop extending $r_t$ and add it to the set of finished routes.

On odd-numbered timesteps, the available actions are drawn from $\textup{SP}$, the set of shortest paths between all pairs of nodes in $\mathcal{G}$.  If $r_t = []$, then:
\begin{equation}
 \mathcal{A}_t = \begin{cases}
    \{a \; | \; a \in \textup{SP}, |a| \leq MAX\} & \text{if} \; r_t = [] \\
    \text{EX}_{r_t} & \text{otherwise}
 \end{cases}
\end{equation}
where $\text{EX}_{r_t}$ is the set of paths $a \in \textup{SP}$ that satisfy all of the following conditions:
\begin{itemize}
    \item $(i,j,\tau_{ij}) \in \mathcal{E}_s$, where $i$ is the first node of $a$ and $j$ is the last node of $r_t$, or vice-versa
    \item $a$ and $r_t$ have no nodes in common
    \item $|a| \leq MAX - |r_t|$
\end{itemize}
Once a path $a_t \in \mathcal{A}_t$ is chosen, $r_{t+1}$ is formed by appending $a_t$ to the beginning or end of $r_t$ as appropriate.

On even-numbered timesteps, the action space depends on the number of stops in $r_t$:
\begin{align}\label{eqn:halt_actions}
    \mathcal{A}_t = \begin{cases}
        \{\textup{continue}\} & \text{if } |r_t| < MIN \\
        \{\textup{halt}\} & \text{if } |r_t| = MAX \text{ or } |\text{EX}_{r_t}| = 0 \\
        \{\textup{continue}, \textup{halt}\} & \text{otherwise}
    \end{cases}
\end{align}

And the route set $\mathcal{R}_{t+1}$ and current route $r_{t+1}$ are determined by the chosen action as follows:
\begin{align}
(\mathcal{R}_{t+1}, r_{t+1}) = \begin{cases}
    (\mathcal{R}_t \cup \{r_t\}, []) & \text{if } a_t = \textup{halt} \\
    (\mathcal{R}_t, r_t) & \text{if } a_t = \textup{continue}
\end{cases}
\end{align}

The episode ends when the $S$-th route is added to $\mathcal{R}$: that is, if $|\mathcal{R}_{t+1}| = S$, the episode ends at timestep $t$, and $\mathcal{R}_{t+1}$ is the final transit network.  The reward function is defined as the decrease in cost between each consecutive pair of timesteps:
\begin{equation}\label{eqn:reward_fn}
R_t = C'(\mathcal{G}, \mathcal{R}_t \cup \{r_t\}) - C'(\mathcal{G}, \mathcal{R}_{t+1} \cup \{r_{t+1}\})
\end{equation}
where $C'((\mathcal{G}, \mathcal{R}) = C(\mathcal{G}, \mathcal{R}) - \beta F_s$; that is, $C'$ is the cost function without the $F_s$ term that penalizes violations of constraint 3.  We remove $F_s$ because the \ac{MDP} structure guarantees that the resulting networks satisfy constraint 3.


This \ac{MDP} formalization imposes some helpful biases on the space of solutions.  First, it requires any route connecting $i$ and $j$ to stop at all nodes along some path between $i$ and $j$, biasing planned routes towards covering more nodes.  Second, it biases routes towards directness by forcing them to be composed of shortest paths.  While an agent may construct indirect routes by choosing paths with length 2 at every step, this is unlikely because in realistic link graphs, the majority of paths in $\textup{SP}$ are longer than two nodes.  Third, the alternation between deciding to continue or halt $r_t$ and deciding how to extend $r_t$ means that the probability of halting does not depend on how many different extensions are available; so a policy learned in environments with few possible extensions should generalize to environments with many, and vice versa.

\section{Neural Heuristics}\label{sec:system}

\subsection{Learning to Construct a Network}

To build a neural heuristic, we first train a \ac{GNN} policy $\pi_\theta(a|s)$, parameterized by $\theta$, to maximize the cumulative return $G_t$ on the construction \ac{MDP} described in \autoref{subsec:mdp}.  By following this learned policy on the \ac{MDP} for some city $\mathcal{G}$, stochastically sampling $a_t$ from $\pi(\cdot|s_t)$ at each timestep, we obtain a transit network $\mathcal{R}$ for that city.  We refer to this learned-policy-based construction algorithm as `learned construction'.  Because learned construction samples actions stochastically, we can run it repeatedly to generate multiple networks for a city.  `LC-100' denotes the procedure whereby we run learned construction 100 times and choose the best of the resulting networks.     
As noted in \autoref{subsec:rw_ptopt}, the \ac{NDP} has a much wider space of possible solutions than problems such as the \ac{TSP} or \ac{VRP}.  To make this tractable, we found it necessary to `factorize' the computation of the policy on odd-numbered timesteps, since $\mathcal{A}_t$ may be very large.  Our policy net outputs a scalar score $o_{aij}$ for each node pair $(i,j)$ in the graph and each possible extension $a$.  From these, a score $o_a$ for each extension is computed by summing over $o_{aij}$ for all $(i,j)$ that would become directly connected by appending path $a$ to the current route $r_t$:
\begin{equation}
    o_{a} = \sum_{i \in a, j \in a, i \neq j} o_{aij} + \sum_{i \in r_t, j \in a} o_{aij}
\end{equation}

where $o_{aii} = 0$ for all $i$.  With this factorization, the learning objective becomes estimating the benefit of directly connecting nodes $i$ and $j$ with travel time $\tau_{(r_t|a)ij}$.  This is an easier objective than directly estimating the merit of adding path $a$ to $r_t$.

The central component of the policy net is a \acf{GAT} \citep{gatv2conv} which treats the city $\mathcal{G}$ as a fully-connected graph with node set $\mathcal{N}$.  Each node $i$ has a feature vector $\mathbf{x}_i$, and each pair of nodes $(i,j)$ also has a feature vector $\mathbf{e}_{ij}$.  These contain information about location, demand, existing transit routes, and the link edge (if one exists) between $i$ and $j$.  

The \ac{GAT} takes the sets of node and node-pair feature vectors, and outputs a set of node embeddings $Y = \{\mathbf{y}_i \; \forall i \in \mathcal{N}\}$.  These are used as input to one of two `head' neural nets, $\textup{NN}_{ext}$ or $\textup{NN}_{halt}$, depending on the timestep.  When the timestep $t$ is odd, $\textup{NN}_{ext}$ takes $Y$ and the driving time from $i$ to $j$ along $r_t | a$ (the concatenation of $r_t$ with $a$), and computes the node-pair scores $o_{ij} \; \forall \; i,j \in \mathcal{N} \times \mathcal{N}$.  When $t$ is even, $\textup{NN}_{halt}$ takes $Y$ and other statistics of $\mathcal{G}$ and $\mathcal{R}_t \cup \{r_t\}$ and outputs a probability of choosing `halt' vs `continue'.  The details of these components are presented in \autoref{apx:arch}, along with the details of the node and edge feature vectors.  We have released the code for our method and experiments to the public\footnote[2]{ Available at \url{https://github.com/aholliday/transit_learning}}.

\subsubsection{Training}\label{sssec:methodology_training}

In our prior work \citep{holliday2023augmenting, holliday2024autonomous}, we trained the policy net using the `REINFORCE with Baseline' method proposed by~\cite{williams1992reinforce}, with a reward function that gave a reward only at the end of the \ac{MDP}.:
\begin{equation}
R_t = \begin{cases} -C(\mathcal{G}, \mathcal{R}_t) & \text{if } t = t_\text{end} \\
0 & \text{if } t < t_\text{end} \end{cases}
\end{equation}
This was inspired by the similar approach taken by~\cite{Kool2019AttentionLT}.  However, sparse reward functions like this one, where $R_t=0$ at most timesteps, convey little information about the effect of each action, making learning more difficult.  So we here use the richer reward function we describe in section~\ref{subsec:mdp}.

REINFORCE has also been superseded by more recent policy gradient learning algorithms.  So in this work we train the policy net using \acl{PPO} with Generalized Advantage Estimation, a policy gradient method proposed by~\cite{schulman2017PPO}  \ac{PPO} was chosen for its relative simplicity and for the fact that it still achieves state-of-the-art performance on many problems.  \ac{PPO} works by running $H$ timesteps of an \ac{MDP} episode (or multiple episodes in parallel), and then computing an `advantage' $A_t$ for each timestep:
\begin{equation}
A_t = \sum^{H-1}_{t'=0} \gamma^{t'} R_{t+t'} + \gamma^H V(s_{t+H}) - V(s_t)
\end{equation}
From this we compute the generalized advantage, $\hat{A}_t$, a weighted sum of $A_t$ over a range of values of the horizon, $H$.  $\hat{A}_t$ weights the updates to the policy, such that actions with a positive $\hat{A}_t$ are made more likely, and those with a negative $\hat{A}_t$ are made less so. 

The value function $V(s_t)$ is itself a small \ac{MLP} neural net that is trained in parallel with the policy net.  It takes the the cost weight parameter $\alpha$ and statistics of the city $\mathcal{G}$ and of the current network $\mathcal{R}_t \cup \{r_t\}$ as input, and is trained to predict the discounted return $G^H_t$:
\begin{equation}
    G^H_t = \sum^{H-1}_{t'=0} \gamma^{t'} R_{t+t'} + \gamma^H V(s_{t+H})
\end{equation}

We train the policy and value nets on a dataset of synthetic cities.  At each iteration, a full \ac{MDP} episode is run in parallel on each city in a `batch' of cities from the dataset.   We sample a different $\alpha \sim [0, 1]$ for each training \ac{MDP} episode, while holding $S, MIN, MAX$, and the constraint weight $\beta$ constant.  The values to which we set these hyperparameters are presented in \autoref{tab:learning_params}.  $C(\mathcal{G}, \mathcal{R}_t \cup \{r_t\})$, $V(s_t)$ and $\hat{A}_t$ are computed for each timestep of each episode in the batch, and the policy and value nets are updated accordingly via \ac{PPO}.


To construct a synthetic city for the training dataset, we first generate its nodes and link network using one of these processes chosen at random:

\begin{itemize}
    \item $4$-nn: Create the node set $\mathcal{N}$ by sampling $n$ 2D points uniformly at random over a square.  Then create $\mathcal{E}_s$ by making link edges between each node $i$ and its four nearest neighbours.
    \item 4-grid: Place $n$ nodes in a rectangular grid as close to square as possible.  Create $\mathcal{E}_s$ by making link edges between all horizontal and vertical neighbours.
    \item 8-grid: The same as 4-grid, but also make link edges between diagonal neighbours.
    \item Voronoi: Sample $m$ random 2D points in a square, and compute their Voronoi diagram~\citep{fortune1995voronoi}.  Let $\mathcal{N}$ and $\mathcal{E}_s$ be the shared vertices and edges of the resulting Voronoi cells.  $m$ is chosen so $|\mathcal{N}| = n$.    
\end{itemize}

In each process, we sample nodes (or the $m$ points for Voronoi) in a $30 \textup{km} \times 30 \textup{km}$ square, and assume a fixed vehicle speed of $v = 15 \textup{m/s}$ to calculate link edge weights $\tau_{ij} = ||(x_i, y_i) - (x_j, y_j)||_2 / v$.  After each process except Voronoi, we delete each edge in $\mathcal{E}_s$ with probability $\rho$:  if the resulting graph $\{\mathcal{N}, \mathcal{E}_s\}$ is not connected, we discard it and repeat the process.  This edge-deletion step adds more variety to the layouts of each type of city, especially 4-grid and 8-grid types, which would otherwise all have the same structure.

Finally, we generate the demand matrix $D$ by setting diagonal demands $D_{ii} = 0$, uniformly sampling above-diagonal elements $D_{ij}$ in the range $[60, 800]$, and then setting below-diagonal elements $D_{ji} = D_{ij} \text{ if } i < j$ to make $D$ symmetric.  

We generate a dataset of $2^{15} = 32,768$ of these synthetic cities, setting $n=20$ and $\rho=0.3$.  During training, we randomly choose 90\% of these cities to form the training set, and set aside the remaining 10\% as a validation set.


In each iteration of training, we train on a batch of cities from the training set.  After every ten iterations, we run learned construction with the current policy net on every city in the validation set, and record the average cost $C_{val}$ of the resulting networks.  At the end of training, we return the parameters $\theta$ that achieved the lowest $C_{val}$, giving the final policy $\pi_\theta$.  Training proceeds for 200 iterations in batches of 256 cities, as we found that the policy stopped improving after this.

During training, all neural net inputs are normalized so as to have zero mean and unit variance across the entire dataset.  The scaling and shifting parameters are saved as part of the policy net and are applied to new data presented to $\pi_\theta$ after training.

\subsection{Evolutionary Algorithm}\label{subsec:evo_alg}


To test the usefulness of a neural net policy for metaheuristic algorithms, we devised a `neural heuristic' that makes use of such a policy, and compared a simple baseline evolutionary algorithm to a variant of the baseline that uses this neural heuristic.  In this section, we describe these two evolutionary algorithms, as well as our neural heuristic.

Our baseline evolutionary algorithm is based on that of~\cite{nikolic2013transit}, with several modifications that we found improved its performance.  We chose this algorithm because of its good performance for relatively little computational cost, but we note that our learned policies could be used as heuristics in a wide variety of metaheuristic algorithms.  \cite{nikolic2013transit} describe their algorithm as a `bee colony optimization' algorithm.  As noted in \cite{sorensen2015metaheuristics}, `bee colony optimization' is mathematically identical to one kind of evolutionary algorithm, an older and well-established class of metaheuristic algorithms.  To avoid confusion and the spread of unnecessary terminology, we here refer to as an evolutionary algorithm and avoid the `bee colony' terminology.

The algorithm operates on a population of $B$ solutions $\mathcal{B} = \{\mathcal{R}_b | 1 \leq b \leq B\}$, and performs alternating stages of mutation and selection.  In the mutation stage, the algorithm applies two heuristics, type 1 and type 2, to different subsets of the population chosen at random.  If the modified network $\mathcal{R}'_b$ has lower cost than its `parent' $\mathcal{R}_b$, it replaces its parent in $\mathcal{B}$: $\mathcal{R}_b \leftarrow \mathcal{R}'_b$.  This is repeated $E$ times in the stage.  Then, in the selection stage, solutions  either `die' or `reproduce', with probabilities inversely related to their cost $C(\mathcal{G}, \mathcal{R}_b)$.  After $IT$ repetitions of mutation and selection, the algorithm returns the network $\mathcal{R}$ with the lowest $C(\mathcal{G}, \mathcal{R})$ from all iterations.

The type-1 and type-2 heuristics both begin by selecting, uniformly at random, a route $r$ in $\mathcal{R}_b$ and a terminal node $i$ on $r$.  The type-1 heuristic then selects a random node $j \neq i$ in $\mathcal{N}$, and replaces $r$ with $\text{SP}_{ij}$, the shortest path between $i$ and $j$.  The probability of choosing each node $j$ is proportional to the amount of demand directly satisfied by $\text{SP}_{ij}$.  The type-2 heuristic does one of two things: with  probability $p_d$, it deletes $i$ from $r$; otherwise, it adds a random node $j$ in $i$'s link-graph neighbourhood to $r$, making $j$ the new terminal.  Following~\cite{nikolic2013transit}, we set $p_d = 0.2$ in our experiments.

The initial population of solutions is obtained by making $B$ copies of a single initial solution, $\mathcal{R}_0$.  In our prior work~\citep{holliday2023augmenting}, we found that using LC-100 to obtain $\mathcal{R}_0$ outperformed \cite{nikolic2013transit}'s own method, so that is what both baseline and variant evolutionary algorithms do here.  We refer to this baseline evolutionary algorithm as `EA'.

\begin{algorithm}
\caption{Evolutionary Algorithm (Baseline and Neural)}\label{alg:ea}
\begin{algorithmic}[1]
\State {\bfseries Input:} $\mathcal{G} = (\mathcal{N}, \mathcal{E}_s, D), \textup{SP}, C, B, IT, E$
\State Construct initial network $\mathcal{R}_0 \leftarrow \text{LC-100}(\mathcal{G}, C)$
\State $\mathcal{R}_b \leftarrow \mathcal{R}_0 \; \forall \; b \in \text{integers } 1 \text{ through } B$
\State $\mathcal{R}_\textup{best} \leftarrow \mathcal{R}_0$
\For {$i=1$ to $IT$}
    \State // Mutation stage
    \For{$j=1$ to $E$}
        \For{$b=1$ to $B$}
            \If {$b \leq B / 2$}
                \If {baseline algorithm (EA)}: $\mathcal{R}'_b \leftarrow$ type\_1\_heuristic($\mathcal{R}_b$)
                \EndIf
                \If {neural variant (NEA)}: $\mathcal{R}'_b \leftarrow$ neural\_heuristic($\mathcal{R}_b$)
                \EndIf
            \Else
                \State $\mathcal{R}'_b \leftarrow$ type\_2\_heuristic($\mathcal{R}_b$)
            \EndIf
            \If {$C(\mathcal{G}, \mathcal{R}'_b) < C(\mathcal{G}, \mathcal{R}_b)$}
                \State $\mathcal{R}_b \leftarrow \mathcal{R}'_b$
            \EndIf
        \EndFor
        \State Randomly shuffle network indices $b$
    \EndFor
    \State // Selection stage
    \State $C_{max} \leftarrow \max\limits_b C(\mathcal{G}, \mathcal{R}_b)$
    \State $C_{min} \leftarrow \min\limits_b C(\mathcal{G}, \mathcal{R}_b)$
    \For{$b=1$ to $B$}
        \If {$C(\mathcal{G}, \mathcal{R}_b) < C(\mathcal{G}, \mathcal{R}_\textup{best})$}
            \State $\mathcal{R}_\textup{best} \leftarrow \mathcal{R}_b$
        \EndIf
        \State // Select surviving networks
        \State $O_b \leftarrow \frac{C_{max} - C_b}{C_{max} - C_{min}}$
        \State $s_b \sim \textup{Bernoulli}(1 - e^{-O_b})$
        \State // Set survivor reproduction probabilities
        \State $p_b \leftarrow \frac{O_bs_b}{\sum_{b'} O_{b'}s_{b'}}$        
    \EndFor
    \If {$\exists \; b \in [1, B]$ s.t. $s_b = 1$}
        \State // Replace non-survivors with survivors' offspring 
        \For{$b=1$ to $B$}
            \If {$s_b = 0$}
                \State $k \sim P(K)$, where $P(K=b') = p_{b'}$
                \State $\mathcal{R}_b \leftarrow \mathcal{R}_k$
            \EndIf
        \EndFor        
    \EndIf
\EndFor
\State return $\mathcal{R}_\textup{best}$
\end{algorithmic}
\end{algorithm}

Our variant algorithm differs from EA only in that instead of using the type-1 heuristic, it uses a `neural heuristic' which is based on our neural net policy.  The neural heuristic selects a route $r \in \mathcal{R}_b$ at random, removes it from the network to get $\hat{\mathcal{R}_b} = \mathcal{R}_b \setminus r$, and then runs learned construction with $\pi_\theta$ starting from $\hat{\mathcal{R}_b}$.  Because $|\hat{\mathcal{R}_b}| = S-1$, this produces just one new route, $r'$, which is then added to the network to get $\mathcal{R}_b' \leftarrow \hat{\mathcal{R}_b} \cup \{ r' \}$.  In this way we leverage $\pi_\theta$, which was trained as a policy for a construction method, to aid in an improvement method.  We refer to this variant algorithm as the neural evolutionary algorithm (NEA).  Both algorithms are presented in \autoref{alg:ea}, with the difference between the two described in lines 10 and 11.

We replace the type-1 heuristic with the neural heuristic because they each choose from a similar space of changes: replacing one route by a shortest path in the former case, and replacing one route by a new route composed of shortest paths in the latter.

\section{Benchmark Experiments}\label{sec:mumford_exp}

We first evaluated our method on the \cite{mandl1980evaluation} and \cite{mumford2013dataset} city datasets, two popular benchmarks for evaluating \ac{NDP} algorithms~\citep{mumford2013new, john2014routing, kilic2014demand, ahmed2019hyperheuristic}.  Mandl is just one small synthetic city with fifteen nodes.  The Mumford dataset consists of four synthetic cities, labeled Mumford0 through Mumford3, and it specifies values of $S$, $MIN$, and $MAX$ to use when benchmarking on each city.  The values $n$, $S$, $MIN$, and $MAX$ for Mumford1, Mumford2, and Mumford3 are taken from three different real-world cities and their existing transit networks, giving the dataset a degree of realism.  More details of these benchmark cities are given in \autoref{tab:dataset}.

\begin{table*}
    \centering
    \caption{Statistics of the Mandl and Mumford benchmark cities.}    
    \resizebox{1.0\textwidth}{!}{
    \label{tab:dataset}
    \begin{tabular}{lcccccc}
        \toprule
          City & \# nodes $n$ & \# link edges $|\mathcal{E}_s|$ & \# routes $S$ & $MIN$ & $MAX$ & Area (km$^2$) \\
          \midrule
          Mandl   & 15  & 20  & 6  & 2  & 8 & 352.7 \\
         Mumford0 & 30  & 90  & 12 & 2  & 15 & 354.2 \\
         Mumford1 & 70  & 210 & 15 & 10 & 30 & 858.5 \\
         Mumford2 & 110 & 385 & 56 & 10 & 22 & 1394.3 \\
         Mumford3 & 127 & 425 & 60 & 12 & 25 & 1703.2 \\
         \bottomrule
    \end{tabular}
    }
\end{table*}

In all of our experiments, we set the transfer time penalty used in computing the average trip time $C_p$ to $300$s (five minutes).  This value 
is widely used in other work that reports results on these \ac{MDP} benchmarks~\citep{mumford2013new}, so using this value of the transfer time penalty allows us to directly compare our own results with these other results.

\subsection{Comparison with Baseline Evolutionary Algorithm}\label{subsec:baseline}

We ran each algorithm under consideration on all five of these synthetic cities over eleven different values of $\alpha$, ranging from $0.0$ to $1.0$ in increments of $0.1$.  This let us observe how well the different methods perform over a range of possible user preferences, including the extremes of the operator perspective ($\alpha = 0.0$, caring only about $C_o$) and passenger perspective ($\alpha = 1.0$, caring only about $C_p$) as well as a range of intermediate perspectives.  In all of these experiments, we set the constraint weight $\beta$ to $5.0$, the same value used in training the policies.  We found that this was sufficient to prevent any of the constraints listed in \autoref{sec:tndp} from being violated by any transit network produced in our experiments.

Because these are stochastic algorithms, for each benchmark city we ran each algorithm ten times with ten different random seeds.  For algorithms that make use of a learned policy, we trained ten separate policies with the same set of random seeds (but using the same training dataset), and used each policy when running algorithms with the corresponding random seed.  The values we report are statistics averaged over these ten runs.

We first compared our neural evolutionary algorithm (NEA) with our baseline evolutionary algorithm (EA).  We also compared both with the initial networks from LC-100, to see how much improvement each algorithm makes over the initial networks.  In the EA and NEA runs, the same parameter settings of $B=10, IT=400, E=10$ were used, following the values used in~\cite{nikolic2013transit}.  

\autoref{tab:baseline} displays the cost $C(\mathcal{G}, \mathcal{R})$ achieved by each algorithm on the Mandl and Mumford benchmarks for the operator perspective ($\alpha=0.0$), the passenger perspective ($\alpha=1.0$), and a balanced perspective ($\alpha=0.5$).  We see that on Mandl, the smallest of the five cities, EA and NEA perform virtually identically.  But for the three largest cities, NEA performs considerably better than EA for both the operator and balanced perspectives.  On Mumford3, the largest of the five cities, NEA solutions have 13\% lower average cost for the operator perspective and 5\% lower for the balanced perspective than EA.  For the passenger perspective, NEA's advantage over EA is smaller, but it still outperforms EA on all of the Mumford cities.  

\begin{table*}[]
    \centering
	\caption{Cost $C(\alpha, \mathcal{G}, \mathcal{R})$ achieved by LC-100, EA, and NEA for three different $\alpha$ values.  Costs are averaged over ten random seeds; the $\pm$ value is the standard deviation of $C(\alpha, \mathcal{G}, \mathcal{R})$ over the seeds.}
    \resizebox{1.0\textwidth}{!}{    
	\begin{tabular}{ccrcccccc}
		\toprule
		$\alpha$ & Method &   Mandl &         Mumford0 &         Mumford1 &         Mumford2 &         Mumford3 \\               
		\midrule 
		& LC-100 & 0.697 $\pm$ 0.011 &  0.847 $\pm$ 0.025 &  1.747 $\pm$ 0.034 &  1.315 $\pm$ 0.049 &  1.333 $\pm$ 0.064 \\
		0.0 & EA & \bf 0.687 $\pm$ 0.016 & \bf 0.776 $\pm$ 0.026 &  1.637 $\pm$ 0.065 &  1.067 $\pm$ 0.032 &  1.052 $\pm$ 0.041 \\
		& NEA    & \bf 0.687 $\pm$ 0.016 &  0.783 $\pm$ 0.027 & \bf 1.347 $\pm$ 0.033 & \bf 0.938 $\pm$ 0.024 & \bf 0.927 $\pm$ 0.014 \\
		\midrule
		& LC-100 &  0.558 $\pm$ 0.003 &  0.916 $\pm$ 0.006 &  1.272 $\pm$ 0.018 &  0.989 $\pm$ 0.021 &  0.984 $\pm$ 0.026 \\
		0.5 & EA     &  \bf 0.549 $\pm$ 0.008 &  0.841 $\pm$ 0.019 &  1.203 $\pm$ 0.026 &  0.866 $\pm$ 0.023 &  0.851 $\pm$ 0.025 \\
		& NEA    &  0.550 $\pm$ 0.006 & \bf 0.840 $\pm$ 0.021 & \bf 1.052 $\pm$ 0.013 &  \bf 0.816 $\pm$ 0.008 & \bf 0.801 $\pm$ 0.008 \\
		\midrule
		& LC-100 &  0.328 $\pm$ 0.001 &  0.721 $\pm$ 0.004 &  0.573 $\pm$ 0.004 &  0.495 $\pm$ 0.002 &  0.476 $\pm$ 0.001 \\
		1.0 & EA &  \bf 0.315 $\pm$ 0.001 &  0.590 $\pm$ 0.005 &  0.523 $\pm$ 0.003 &  0.480 $\pm$ 0.001 &  0.464 $\pm$ 0.002 \\
		& NEA    &  \bf 0.315 $\pm$ 0.001 & \bf 0.587 $\pm$ 0.004 & \bf 0.520 $\pm$ 0.004 & \bf 0.479 $\pm$ 0.002 &  \bf 0.460 $\pm$ 0.003 \\		
		\bottomrule
	\end{tabular}
    }
    \label{tab:baseline}
\end{table*}

\begin{figure*}
    \centering
    \includegraphics[width=0.9\textwidth]{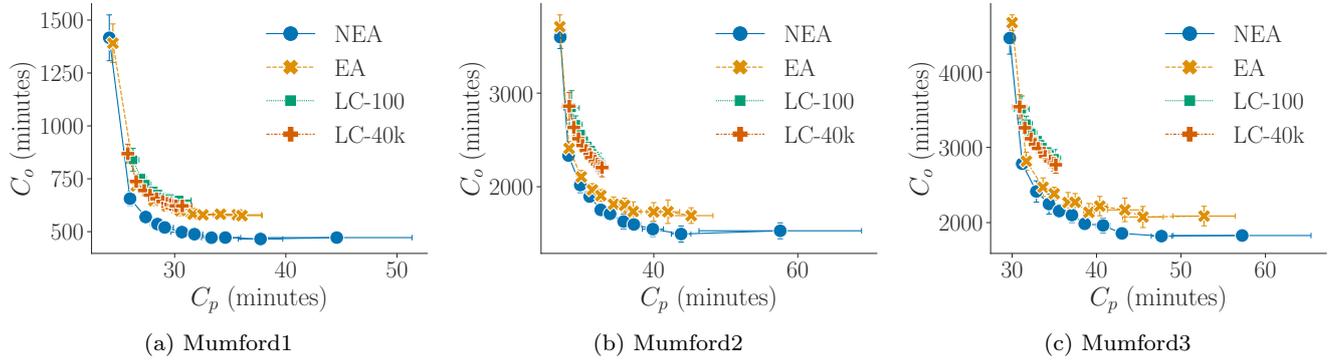}
    \caption{Trade-offs between average trip time $C_p$ (on the x-axis) and total route time $C_o$ (on the y-axis), across values of $\alpha$ over the range $[0, 1]$ in increments of $0.1$, from $\alpha=0.0$ at lower-right to $1.0$ at upper-left, for transit networks from LC-100, LC-40k, EA, and NEA.  Each point shows the mean $C_p$ and $C_o$ over 10 random seeds for one value of $\alpha$, and bars around each point indicate one standard deviation on each axis.  Lines linking pairs of points indicate that they represent consecutive $\alpha$ values.  Lower values of $C_o$ and $C_p$ are better, so the down-and-leftward direction in each plot represents improvement.}
    \label{fig:40k}
\end{figure*}

\autoref{fig:40k} displays the average trip time $C_p$ and total route time $C_o$ achieved by each algorithm on the three largest Mumford cities, each of which are based on a real city's statistics, at eleven $\alpha$ values evenly spaced over the range $[0, 1]$ in increments of 0.1.  We observe that there is a trade-off between $C_p$ and $C_o$: $C_o$ increases and $C_p$ decreases as $\alpha$ increases.  We also observe that for $\alpha < 1.0$, NEA pushes $C_o$ much lower than LC-100 and EA.  Importantly, for any network $\mathcal{R}$ from LC-100 or EA, there is some value of $\alpha$ for which NEA produces a network $\mathcal{R}'$ that strictly dominates $\mathcal{R}$, having equal or lower $C_p$ and $C_o$.  \autoref{fig:40k} also shows results for an additional algorithm, LC-40k, which we discuss in \autoref{subsec:ablations}.

On Mumford1, Mumford2, and Mumford3, we observe a common pattern: EA and NEA perform very similarly at $\alpha=1.0$ (the leftmost point on each curve), but a significant performance gap forms as $\alpha$ decreases.  We also observe that LC-100 favours reducing $C_p$ over $C_o$, with its points clustered at higher $C_o$ and lower $C_p$ than the points with corresponding $\alpha$ values on the other curves.  EA and NEA each achieve only very small decreases in $C_p$ on LC-100's initial solutions at $\alpha=1.0$, but much larger decreases in $C_o$ at lower $\alpha$.

\begin{figure*}
    \centering
    \includegraphics[width=0.9\textwidth]{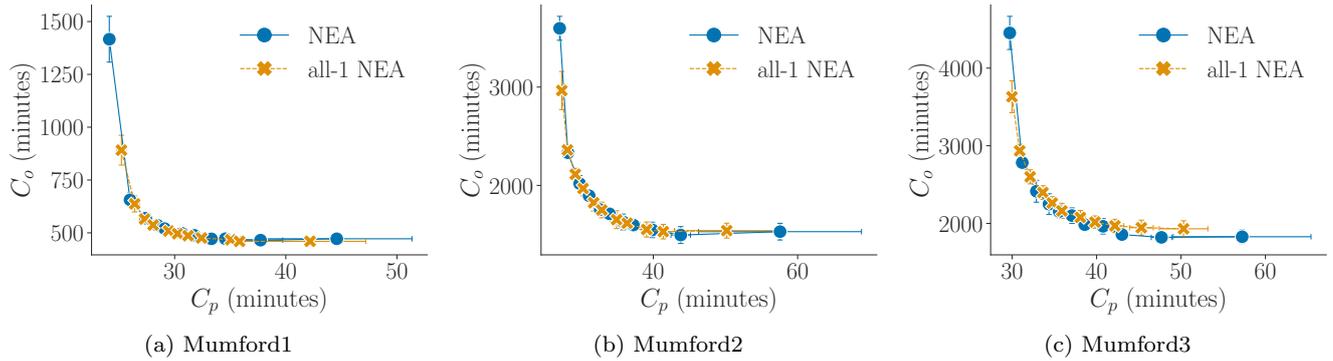}
    \caption{Trade-offs between average trip time $C_p$ (on the x-axis) and total route time $C_o$ (on the y-axis) achieved by all-1 NEA, plotted along with NEA (repeated from \autoref{fig:40k}) for comparison.}
    \label{fig:no2}
\end{figure*}

\begin{figure*}
    \centering
    \includegraphics[width=0.9\textwidth]{figs/rcea.jpg}
    \caption{Trade-offs between average trip time $C_p$ (on the x-axis) and total route time $C_o$ (on the y-axis) achieved by RC-EA, plotted along with NEA (repeated from \autoref{fig:40k}) for comparison.}
    \label{fig:random}
\end{figure*}

\subsection{Ablation Studies}\label{subsec:ablations}

To better understand the contribution of various components of our method, we performed three sets of ablation studies.  These were conducted over the three largest Mumford cities (1, 2, and 3), and over the same range of $\alpha$ values as in \autoref{subsec:baseline}.  

\subsubsection{Effect of number of samples}

Over the course of the evolutionary algorithm with our parameter settings $(B=10, IT=400, E=10)$, a total of $B \times IT \times E = 40,000$ different transit networks are considered.  By comparison, LC-100 only considers 100 networks.  To test how much of NEA's superiority to LC-100 is due to this difference, we repeated our experiments with the LC-40k algorithm, which is the same as LC-100 except that it samples $40,000$ networks from the learned-construction procedure instead of $100$.  

Comparing LC-100 and LC-40k in \autoref{fig:40k}, we see that across all three cities and all values of $\alpha$, LC-40k improves on LC-100 only slightly by comparison with EA or NEA.  From this we conclude that the number of networks considered is not the main cause of EA's and NEA's advantage over LC-100: the evolutionary algorithm itself makes most of the difference.

\subsubsection{Effect of type-2 heuristic}

The type-2 heuristic is the same between EA and NEA and is not a learned function.  To understand how important this heuristic is to NEA's performance, we ran `all-1 NEA', a variant of NEA in which the type-2 heuristic is dropped and only the neural heuristic is used.  The results are shown in \autoref{fig:no2}, along with the curve for NEA (reproduced from \autoref{fig:40k}) for comparison.  

It is clear that NEA and all-1 NEA perform very similarly in most scenarios.  The main difference is at $\alpha=1.0$, where we see that all-1 NEA under-performs NEA, achieving higher $C_p$ (though it also achieves lower $C_o$, due to the inherent trade-off between $C_p$ and $C_o$).  It appears that the type-2 heuristic's ability to extend existing routes is important at $\alpha=1.0$, where making routes longer cannot make the network worse, but otherwise at $\alpha < 1.0$ the neural heuristic does just as well on its own, without the type-2 heuristic.

\subsubsection{Effect of reinforcement learning}\label{sssec:rcea}

The type-1 heuristic used in EA and the neural heuristic used in NEA differ in the kind of routes they create.  The type-1 heuristic can only add routes that are shortest paths in the link graph, while the neural heuristic may compose multiple shortest paths into a new route.  We sought to determine how much of NEA's advantage over EA is due to this structural difference, versus being due to the specific policy $\pi_\theta$ that we trained using \ac{RL}.

To answer this question, we ran a variant of NEA in which $\pi_\theta$ is replaced by a purely random policy $\pi_\text{random}$: 
\begin{equation}
\pi_\text{random}(a | s_t) = \frac{1}{|\mathcal{A}_t|} \; \forall \; a \in \mathcal{A}_t
\end{equation}
We call this variant of NEA the randomly-combining evolutionary algorithm (RC-EA).  In order to fully isolate RC-EA from the effects of \ac{RL}, we also used $\pi_\text{random}$ in the LC-100 algorithm to generate the starting network $\mathcal{R}_0$ for RC-EA.

The results are shown in \autoref{fig:random}, again with the same NEA curve from \autoref{fig:40k} for comparison.  For values of $\alpha < 1.0$, the transit networks from RC-EA are dominated by some transit network from NEA.  But at the extreme of $\alpha = 1.0$, RC-EA's network performs better than any of NEA's networks in terms of average trip time $C_p$, decreasing it by about twenty seconds versus NEA's lowest-$C_p$ network on each city - an improvement of about 1\%.

RC-EA's poor performance at $\alpha < 1$ makes sense: the ability to replace routes with composites of shortest paths biases it towards making longer routes, and unlike $\pi_\theta$, $\pi_\text{random}$ cannot learn to prefer halting early in such cases.  This is a clear disadvantage at $\alpha < 1.0$.  The advantage of NEA comes mainly from what the policy $\pi_\theta$ learns during training.  It is surprising, though, that RC-EA outperforms NEA at $\alpha=1.0$.  We would expect that the neural net policy used in NEA should be able to equal the performance of RC-EA by learning to choose actions uniformly at random when $\alpha=1.0$, yet here it fails to do so.

Nonetheless, NEA performs better overall than RC-EA, and so we leave answering this question to future work, while noting that composing shortest paths randomly is a good heuristic for minimizing $C_p$ at the expense of $C_o$, if that is one's aim.

\subsection{Comparisons with prior work}\label{subsec:vs_prior}

We next sought to compare NEA directly against other \ac{NDP} algorithms from the literature.  It is common practice in this literature to report results on the Mandl and Mumford benchmarks only for the operator perspective ($\alpha=0.0$) and the passenger perspective ($\alpha=1.0$), so in this section we compare NEA against other algorithms only on these two settings of $\alpha$.

While conducting the experiments of \autoref{subsec:baseline}, we observed that after 400 iterations the cost of NEA's best-so-far network $\mathcal{R}_\text{best}$ was still decreasing from one iteration to the next.  We thus performed a further set of experiments where we ran NEA with $IT=4,000$ instead of $IT=400$.  In addition, because we observed good results from RC-EA at $\alpha = 1.0$, we ran RC-EA with $IT=4,000$ at $\alpha = 1.0$, and we include these results in the comparison below.  On a desktop computer with a 2.4 GHz Intel i9-12900F processor and an NVIDIA RTX 3090 graphics processing unit (used to accelerate our neural net computations), NEA takes about 10 hours for each 4,000-iteration run on Mumford3, the largest environment, while RC-EA takes about 6 hours.

By comparison, on Mumford3, \cite{mumford2013new} report no running times; \cite{zervas2024solving} run their algorithm for 3 hours; \cite{kilic2014demand} do not report total running time, but take 8 hours just to construct the initial network; \cite{ahmed2019hyperheuristic} run their algorithm for 10 hours; and \cite{john2014routing} and \cite{husselmann2023improved} run theirs for more than two days.  So NEA with $IT=4,000$ falls near the middle of the distribution of computational costs of methods we compare it with, making this a fair comparison.

\autoref{tab:benchmark_pp} and \autoref{tab:benchmark_op} present the results of these $IT=4,000$ runs on the Mandl and Mumford benchmarks, alongside results reported on these benchmarks in comparable recent work.  As elsewhere, the results we report for NEA and RC-EA are averaged over ten runs with different random seeds; the same set of ten trained policies used in the experiments of \autoref{subsec:baseline} and \autoref{subsec:ablations} are used here in NEA.  In addition to the average trip time $C_p$ and total route time $C_o$, these tables present $d_0$, $d_1$, $d_2$, and $d_{un}$, which indicate how many transfers between routes passengers had to make when using the transit network: $d_i$ with $i \in \{0, 1, 2\}$ is the percentage of all passenger trips that required $i$ transfers, while $d_{un}$ is the percentage of trips that require more than two transfers: 
\begin{equation}
d_{un} = 100 - (d_0 + d_1 + d_2)
\end{equation}

\subsubsection{Note on omitted work}

In these comparisons, we exclude two recent papers that report results on the Mumford benchmark: these are \cite{islam2019heuristic}, and \cite{wens2024decentralised}.  Both works report passenger-perspective results that are computed in non-standard ways: \cite{islam2019heuristic} ignore passenger trips that take more than 2 transfers when computing $C_p$, while \cite{wens2024decentralised} ignore constraint 1 from the constraints listed in \autoref{sec:tndp}, allowing the network to fail to connect some pairs of nodes.  This means that their reported $C_p$ values are not directly comparable with our own, nor with those reported in other work on the \ac{NDP}.   We exclude them so that the differences between compared results are attributable only to the algorithms that produced them.

\begin{table*}
\centering
\caption{Passenger-perspective results.  $C_p$ is the average passenger trip time. $C_o$ is the total route time.  $d_i$ is the percentage of trips satisfied with number of transfers $i$, while $d_{un}$ is the percentage of trips satisfied with 3 or more transfers.  Arrows next to each quantity indicate which of increase or decrease is desirable.  Bold values in $C_p$, $d_0$, and $d_{un}$ columns are the best for that city.}
\resizebox{1.0\textwidth}{!}{
\begin{tabular}{llllllll}
\toprule
City & Method & $C_p \downarrow$ & $C_o$ &  $d_0 \uparrow$ &  $d_1$ &  $d_2$ & $d_{un} \downarrow$ \\
\midrule
Mandl & \cite{mumford2013new} &  10.27 &   221 &  95.38 &   4.56 &   0.06 & \bf 0 \\
& \cite{john2014routing} &  10.25 &   212 &      - &      - &      - &        - \\
& \cite{kilic2014demand} &  10.29 &   216 &   95.5 &    4.5 &      0 &        \bf 0 \\
& \cite{ahmed2019hyperheuristic} & \bf 10.18  & 212  & 97.17  & 2.82   & 0.00   & \bf 0 \\
& \cite{husselmann2023improved} DBMOSA &  10.27 &   179 &  95.94 &   3.93 &   0.13 & \bf 0 \\
& \cite{husselmann2023improved} NSGA-II &  10.19 &   197 &  97.36 &   2.64 &      0 & \bf 0 \\
& \cite{zervas2024solving} & 10.19 & - & \bf 97.94 & 2.06 & 0 & \bf 0 \\
 & NEA & 10.37 & 181 & 93.89 & 5.93 & 0.18 & \bf 0 \\
 & RC-EA & 10.27 & 193 & 95.83 & 4.16 & 0.01 & \bf 0 \\

\midrule
Mumford0 & \cite{mumford2013new} &  16.05 &   759 &   63.2 &  35.82 &   0.98 & \bf 0 \\
& \cite{john2014routing} &   15.4 &   745 &      - &      - &      - &  - \\
& \cite{kilic2014demand} &  14.99 &   707 &  69.73 &  30.03 &   0.24 & \bf 0 \\
& \cite{ahmed2019hyperheuristic} & \bf 14.09  & 722  & \bf 88.74  & 11.25   & 0   & \bf 0  \\
& \cite{husselmann2023improved} DBMOSA &  15.48 &   431 &   65.5 &   34.5 & 0 & \bf 0 \\
& \cite{husselmann2023improved} NSGA-II & 14.34 &   635 & 86.94 &  13.06 & 0 & \bf 0 \\
& \cite{zervas2024solving} & 14.76 & - & 91.0 & 9.0 & 0 & 0 \\
& NEA & 15.26 & 639 & 68.35 & 31.24 & 0.41 & \bf 0 \\
& RC-EA & 14.62 & 735 & 77.87 & 22.13 & 0.00 & \bf 0 \\
\midrule
Mumford1 & \cite{mumford2013new} &  24.79 &  2038 &   36.6 &  52.42 &  10.71 &     0.26 \\
& \cite{john2014routing} &  23.91 &  1861 &      - &      - &      - &        - \\
& \cite{kilic2014demand} &  23.25 &  1956 &   45.1 &  49.08 &   5.76 &     0.06 \\
& \cite{ahmed2019hyperheuristic} & \bf 21.69  & 1956  & \bf 65.75  & 34.18   & 0.07   & \bf 0  \\
& \cite{husselmann2023improved} DBMOSA &  22.31 &  1359 &  57.14 &  42.63 &   0.23 & \bf 0 \\
& \cite{husselmann2023improved} NSGA-II & 21.94 &  1851 &  62.11 &  37.84 &   0.05 & \bf 0 \\
& \cite{zervas2024solving} & 22.73 & - & 55.26 & 43.58 & 1.16 & \bf 0 \\
& NEA & 22.85 & 1723 & 49.28 & 48.94 & 1.78 & \bf 0 \\
& RC-EA & 22.29 & 2032 & 53.47 & 45.96 & 0.57 & \bf 0 \\
\midrule
Mumford2 & \cite{mumford2013new} &  28.65 &  5632 &  30.92 &  51.29 &  16.36 &     1.44 \\
& \cite{john2014routing} &  27.02 &  5461 &      - &      - &      - &        - \\
& \cite{kilic2014demand} &  26.82 &  5027 &  33.88 &  57.18 &   8.77 &     0.17 \\
& \cite{ahmed2019hyperheuristic} & 25.19  & 5257  & \bf 56.68  & 43.26   & 0.05   & \bf 0 \\
& \cite{husselmann2023improved} DBMOSA &  25.65 &  3583 &  48.07 &  51.29 &   0.64 & \bf 0 \\
& \cite{husselmann2023improved} NSGA-II & 25.31 &  4171 &  52.56 &  47.33 &   0.11 & \bf 0 \\
& \cite{zervas2024solving} & 25.71 & - & 57.84 & 42.0 & 0.16 & 0 \\
& NEA & 25.25 & 4937 & 52.66 & 47.05 & 0.29 & \bf 0 \\
 & RC-EA & \bf 24.92 & 5655 & 54.12 & 45.85 & 0.04 & \bf 0 \\
\midrule
Mumford3 & \cite{mumford2013new} &  31.44 &  6665 &  27.46 &  50.97 &  18.79 &     2.81 \\
& \cite{john2014routing} &   29.5 &  6320 &      - &      - &      - &        - \\
& \cite{kilic2014demand} &  30.41 &  5834 &  27.56 &  53.25 &  17.51 &     1.68 \\
& \cite{ahmed2019hyperheuristic} & 28.05  & 6119  & 50.41  & 48.81   & 0.77   & \bf 0 \\
& \cite{husselmann2023improved} DBMOSA &  28.22 &  4060 &  45.07 &  54.37 &   0.56 & \bf 0 \\
& \cite{husselmann2023improved} NSGA-II & 28.03 &  5018 &  48.71 &   51.1 &   0.19 & \bf 0 \\
& \cite{zervas2024solving} & 28.1 & - & \bf 61.59 & 38.17 & 0.24 & 0 \\
& NEA & 27.96 & 6127 & 49.36 & 50.04 & 0.60 & \bf 0 \\
& RC-EA & \bf 27.60 & 6896 & 50.90 & 48.99 & 0.11 & \bf 0 \\
\bottomrule
\end{tabular}
}
\label{tab:benchmark_pp}
\end{table*}

\subsubsection{Results}

\autoref{tab:benchmark_pp} shows the passenger perspective results.  On Mandl, Mumford0, and Mumford1, \cite{ahmed2019hyperheuristic}'s method performs best.  But on Mumford3 - the most challenging benchmark city - our NEA algorithm outperforms all other methods in the literature, improving on the previous best of \cite{husselmann2023improved}'s NSGA-II by 0.3\%.  Meanwhile, RC-EA performs best of all on both Mumford2 and Mumford3, improving on Mumford3's previous best $C_p$ by 1.5\%, and setting a new state-of-the-art on these two challenging benchmarks.

\autoref{tab:benchmark_op} shows the operator-perspective results.  \cite{kilic2014demand} and \cite{zervas2024solving} do not report results for the operator perspective, so are not included here.  Similarly to the passenger perspective, NEA here underperforms on the smallest cities (Mandl and Mumford0), but outperforms all methods other than \cite{ahmed2019hyperheuristic} on Mumford1 and Mumford2, including the more recent work of \cite{husselmann2023improved}.  And on the most challenging city, Mumford3, NEA again outperforms all other methods, improving on the previous state-of-the-art $C_o$ by 4.8\%.  

Despite NEA and RC-EA's relatively simple metaheuristic, these two algorithms set a new state of the art on the most challenging of \ac{NDP} benchmarks shows that on large city graphs, our neural heuristic is superior to earlier non-neural heuristics.  And in the passenger-perspective case, our non-neural heuristic of randomly concatenating shortest paths performs even better.  The pattern in these results, where the relative performance of our heuristics increases with the size of the city $n$, also suggests that their advantage may continue to grow for cities larger than those found in these benchmarks.

When optimizing just one metric as in the above experiments, our method achieves low values of that metric by driving the other metric very high, as can be seen in the $C_o$ column of \autoref{tab:benchmark_pp} and the $C_p$ column of \autoref{tab:benchmark_op}.  This is because of the inherent trade-off between $C_p$ and $C_o$: shorter, more direct routes reduce operating cost but connect fewer pairs of nodes directly, requiring passengers to make more transfers and increasing the cost to passengers.  

\begin{table*}
\centering
\caption{Operator perspective results.  $C_p$ is the average passenger trip time. $C_o$ is the total route time.  $d_i$ is the percentage of trips satisfied with number of transfers $i$, while $d_{un}$ is the percentage of trips satsified with 3 or more transfers.  Arrows next to each quantity indicate which of increase or decrease is desirable.  Bold values in $C_o$, $d_0$ and $d_{un}$ columns are the best for that city.}
\resizebox{1.0\textwidth}{!}{
\begin{tabular}{llllllll}
\toprule
	City & Method & $C_o \downarrow$ & $C_p$ & $d_0 \uparrow$ & $d_1$ & $d_2$ & $d_{un} \downarrow$ \\
\midrule
Mandl & \cite{mumford2013new} & \bf 63 & 15.13 & 70.91 & 25.5 & 2.95 & 0.64 \\
& \cite{john2014routing} & \bf 63 & 13.48 & - & - & - & - \\
& \cite{ahmed2019hyperheuristic} & \bf 63 & 14.28  &  62.23 & 27.16  & 9.57  & 1.03 \\
& \cite{husselmann2023improved} DBMOSA & \bf 63 & 13.55 & 70.99 & 24.44 & 4.00 & \bf 0.58 \\
& \cite{husselmann2023improved} NSGA-II & \bf 63 & 13.49 & \bf 71.18 & 25.21 & 2.97 & 0.64 \\
& NEA & 68 & 13.89 & 57.87 & 33.77 & 8.20 & 0.15 \\
\midrule
Mumford0 & \cite{mumford2013new} & 111 & 32.4 & 18.42 & 23.4 & 20.78 & 37.40 \\
& \cite{john2014routing} & 95 & 32.78 & - & - & - & - \\
& \cite{ahmed2019hyperheuristic} & \bf 94  & 26.32  & 14.61  & 31.59   & 36.41  & 17.37 \\
& \cite{husselmann2023improved} DBMOSA & 98 & 27.61 & 22.39 & 31.27 & 18.82 & 27.51 \\
& \cite{husselmann2023improved} NSGA-II & \bf 94 & 27.17 & \bf 24.71 & 38.31 & 26.77 & \bf 10.22 \\
& NEA & 122 & 30.20 & 15.34 & 30.77 & 27.82 & 26.08 \\
\midrule
Mumford1 & \cite{mumford2013new} & 568 & 34.69 & 16.53 & 29.06 & 29.93 & 24.66 \\
& \cite{john2014routing} & 462 & 39.98 & - & - & - & - \\
& \cite{ahmed2019hyperheuristic} & \bf 408  & 39.45  & 18.02  & 29.88  & 31.9  & 20.19 \\
& \cite{husselmann2023improved} DBMOSA & 511 & 26.48 & \bf 25.17 & 59.33 & 14.54 & \bf 0.96 \\
& \cite{husselmann2023improved} NSGA-II & 465 & 31.26 & 19.70 & 42.09 & 33.87 & 4.33 \\
& NEA & 434 & 47.07 & 17.24 & 26.85 & 26.45 & 29.46  \\
\midrule
Mumford2 & \cite{mumford2013new} & 2244 & 36.54 & 13.76 & 27.69 & 29.53 & 29.02 \\
& \cite{john2014routing} & 1875 & 32.33 & - & - & - & - \\
& \cite{ahmed2019hyperheuristic} & \bf 1330  & 46.86  & 13.63  & 23.58   & 23.94  & 38.82 \\ 
& \cite{husselmann2023improved} DBMOSA & 1979 & 29.91 & \bf 22.77 & 58.65 & 18.01 & \bf 0.57 \\
& \cite{husselmann2023improved} NSGA-II & 1545 & 37.52 & 13.48 & 36.79 & 34.33 & 15.39 \\
& NEA & 1356 & 54.82 & 10.99 & 22.41 & 27.00 & 39.60  \\
\midrule
Mumford3 & \cite{mumford2013new} & 2830 & 36.92 & 16.71 & 33.69 & 33.69 & 20.42 \\
& \cite{john2014routing} & 2301 & 36.12 & - & - & - & - \\
& \cite{ahmed2019hyperheuristic} & 1746  & 46.05  & 16.28  & 24.87   & 26.34  & 32.44 \\      
& \cite{husselmann2023improved} DBMOSA & 2682 & 32.33 & \bf 23.55 & 58.05 & 17.18 & \bf 1.23 \\
& \cite{husselmann2023improved} NSGA-II & 2043 & 35.97 & 15.02 & 48.66 & 31.83 & 4.49 \\
& NEA & \bf 1663 & 61.25 & 11.48 & 19.86 & 25.07 & 43.59 \\
\bottomrule
\end{tabular}
}
\label{tab:benchmark_op}
\end{table*}

The metrics $d_0$ to $d_{un}$ reveal that at $\alpha=0.0$, NEA favours higher numbers of transfers relative to most of the other methods, particularly \cite{husselmann2023improved}.  This matches our intuitions: shorter routes will deliver fewer passengers directly to their destinations and will require more transfers, so it is logical that the learned policies would tend to increase the number of transfers when rewarded for minimizing operating cost.  Meanwhile, at $\alpha=1.0$, the numbers of transfers achieved by NEA and RC-EA are comparable to other recent methods, except for \cite{zervas2024solving}.  Unlike our method and the others compared here, their algorithm directly incorporates the number of transfers in the cost function that it minimizes, resulting in fewer transfers but less impressive $C_p$.

\section{Case study: Laval}

\begin{figure*}[]
  \centering
  \begin{subfigure}[t]{0.45\columnwidth} 
    \includegraphics[width=\linewidth]{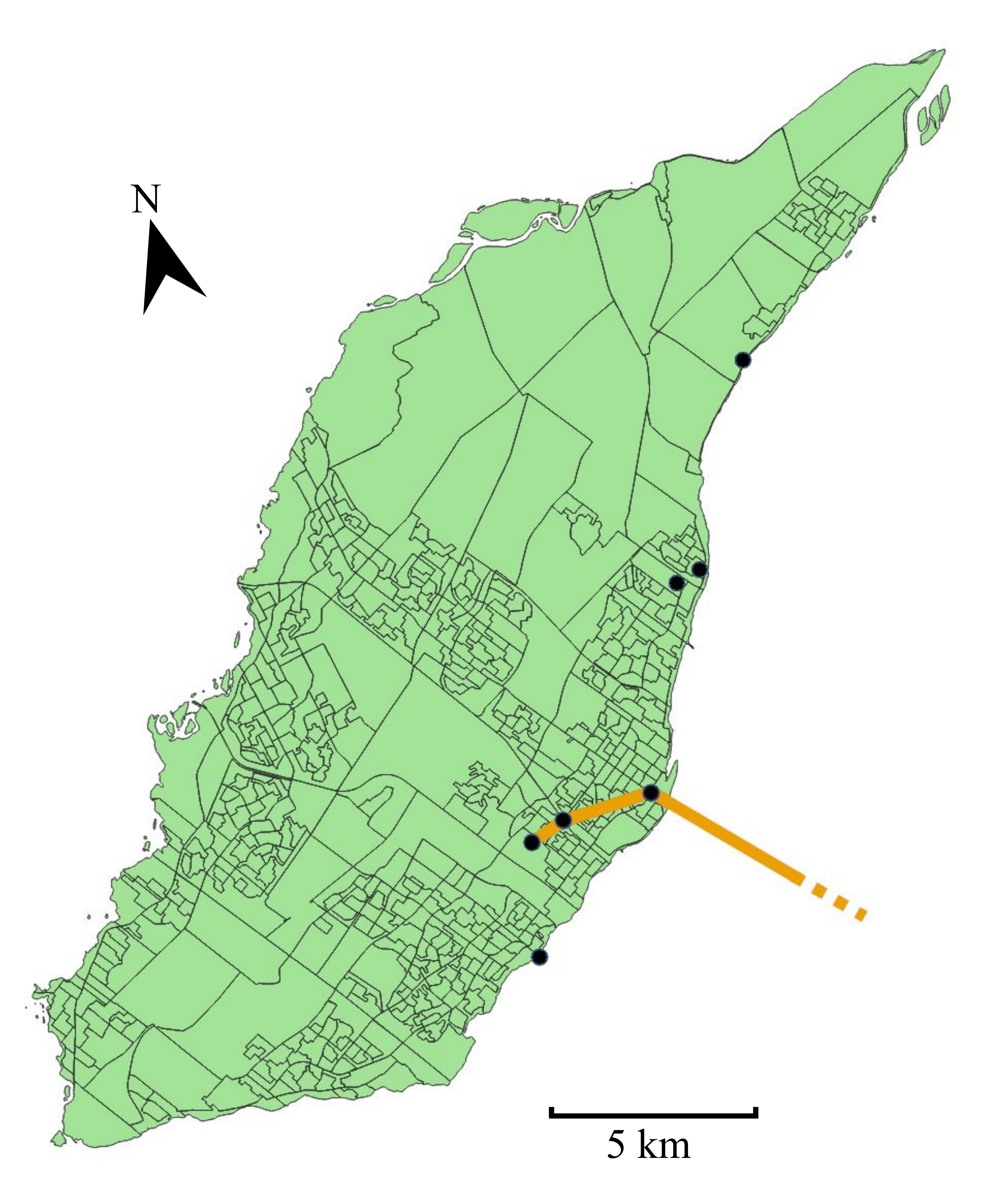}  
    \caption{Census dissemination areas}
    \label{subfig:cda_map}
  \end{subfigure}
  \hfill
  \begin{subfigure}[t]{0.45\columnwidth} 
    \includegraphics[width=\linewidth]{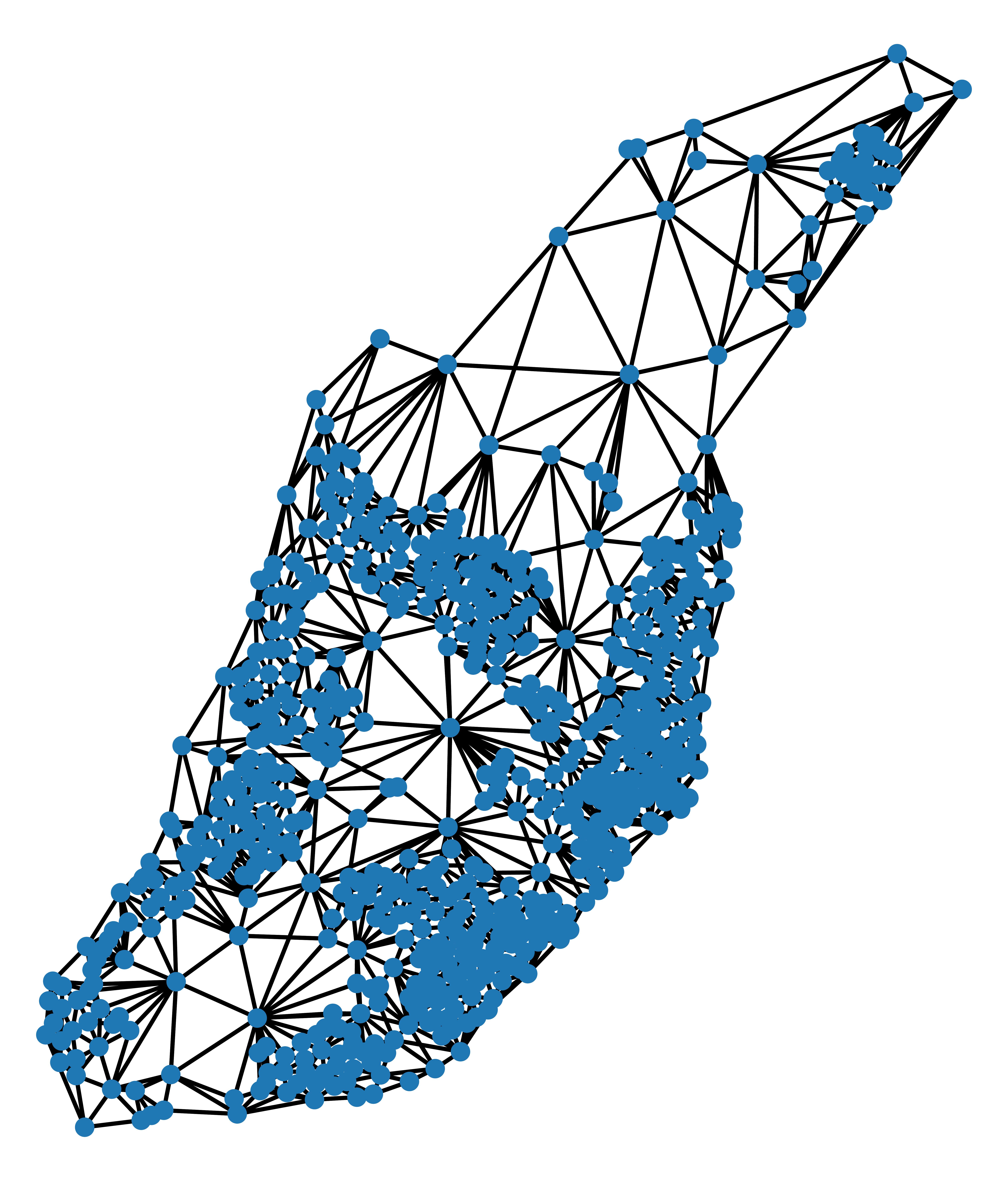}  
    \caption{Link graph}
    \label{subfig:laval_graph}
  \end{subfigure}
  \caption{\autoref{subfig:cda_map} shows a map of the \aclp{CDA} of the city of Laval, with `crossover points' used to remap inter-city demands shown as black circles, and the Montreal Metro Orange Line shown in orange.  \autoref{subfig:laval_graph} shows the link graph constructed from these dissemination areas.}
  \label{fig:laval_maps}
\end{figure*}

To assess our method in a realistic case, we applied it to a simulation of the city of Laval in Quebec, Canada.  Laval is a suburb of the major Canadian city of Montreal (\autoref{fig:laval_maps}).  As of the 2021 census, it had a total population of about 430,000~\citep{canadaCensus2021}.  Public transit in Laval is provided primarily by the bus network of the Soci\'et\'e de Transport de Laval; additionally, the Orange Line of the Montreal underground metro system has three stops in Laval.

We constructed a city graph to represent Laval based on the 2021 \acp{CDA} obtained from Statistics Canada~\citep{disseminationAreas}, a GIS representation of the road network of Laval provided to us by the Soci\'et\'e de Transport de Laval, an \ac{OD} dataset~\citep{od_data} provided by Quebec's Agence M\'etropolitaine de Transport, and publicly-available GTFS data from 2013~\citep{laval_gtfs} that describes Laval's existing transit system.  

Each node in the graph represents the centroid of a \ac{CDA}, with edges connecting it to the nodes of neighbouring \ac{CDA}s, weighted by the average travel time between them over the road network.  The demand matrix $D$ was derived from the \ac{OD} dataset, which gave an estimated number of trips between every pair of locations based on a large survey of Laval residents.  Trips that went outside of Laval to Montreal were redirected to one of several points in the graph where a metro route or bus route in STL's existing transit system crosses into Montreal.  This yielded a graph with 632 nodes.  \autoref{tab:laval_stats} contains the statistics of, and parameters used for, the Laval scenario.

For the full details of how the graph was constructed from this data, we refer the reader to chapter 5 of ~\cite{holliday2024thesis}.

\subsubsection{Existing transit}

Laval's existing transit network has 43 daytime bus routes.  To compare networks from our algorithm to this network, we mapped each stop on these routes to the nearest node in the node set $\mathcal{N}$ based on which \ac{CDA} contains the stop; duplicate stops on the resulting routes were merged.  We refer to this translation of the real transit network as the STL network.  

The Montreal metro system's Orange Line has three stops in Laval. As with the bus routes, we mapped these stops to their containing \acp{CDA} to form an extra metro route $m$, which is treated as being present in all simulations.   We also included this route in the \ac{MDP} by setting $\mathcal{R}_0 = \{m\}$, so that our neural policies would take it as input and account for it.

The numbers of stops on the routes in the STL network range from 2 to 52.  To ensure a fair comparison, we set the constraint parameters to $S=44$ (43 plus 1 for the metro line), $MIN=2$ and $MAX=52$ when running our algorithm.  Unlike the routes in our algorithm's networks, some of STL's routes are asymmetric or unidirectional.  We allowed this in our simulation of STL, while still requiring symmetry and bidirectionality for the transit networks from our algorithms.  For this reason, when reporting total route time $C_o$ in this section, it is calculated by summing the travel times of bidirectional routes in both directions, instead of just one direction as we do elsewhere. 

\begin{table*}[]
    \centering
    \caption{Statistics and parameters of the Laval scenario}    
    \resizebox{1.0\textwidth}{!}{    
    \begin{tabular}{c c c c c c c}
        \# nodes $n$ & \# link edges $|\mathcal{E}_s|$ & \# demand trips & \# routes $S$ & $MIN$ & $MAX$ & Area (km$^2$) \\
        \midrule
        632 & 4,544 & 548,159 & 43 & 2 & 52 & 520.1 \\
    \end{tabular}
    }
    \label{tab:laval_stats}
\end{table*}

\subsection{Enforcing constraint satisfaction}

In the experiments on the Mandl and Mumford benchmark cities described in \autoref{sec:mumford_exp}, the EA and NEA algorithms reliably produced networks that satisfied all of the constraints outlined in \autoref{sec:tndp}.  However, Laval's city graph, with 632 nodes, is considerably larger than the largest Mumford city, which has only 127 nodes.  In our initial experiments on Laval, we found that both EA and NEA consistently failed to satisfy constraint 1, producing networks that left some pairs of nodes disconnected. 

To remedy this, we modified the \ac{MDP} described in \autoref{subsec:mdp} to enforce reduction of unsatisfied demand wherever possible.  Let $c_1(\mathcal{G}, \mathcal{R})$ be the number of node-pairs $i,j$ with $D_{ij} > 0$ for which network $\mathcal{R}$ provides no transit path, and let $\mathcal{R}'_t = \mathcal{R}_t \cup \{ r_t \}$ be the network formed from the finished routes and the in-progress route $r_t$.  We made the following changes to the action space $\mathcal{A}_t$ whenever $c_1(\mathcal{G}, \mathcal{R}'_t) > 0$:  
\begin{itemize}
    \item When the timestep $t$ is even and $\mathcal{A}_t = \{\textup{continue}, \textup{halt}\}$, the `$\textup{halt}$' action is removed from $\mathcal{A}_t$.  This means that the current route must be extended if it is possible to do so without violating another constraint.
    \item When $t$ is odd, if $\mathcal{A}_t$ contains any paths that would reduce $c_1(\mathcal{G}, \mathcal{R}'_t)$ if added to $r_t$, then all paths that would {\it not} reduce $c_1(\mathcal{G}, \mathcal{R}'_t)$ are removed from $\mathcal{A}_t$.  This means that if it is {\it possible} to meet some unmet demand by extending $r_t$, then it {\it must} be done.
\end{itemize}

With these changes to the \ac{MDP}, we found that both LC-100 and NEA reliably produced networks that satisfied all constraints, even using policies $\pi_\theta$ trained without these changes.  The results reported in this section were obtained using this variant of the \ac{MDP}.

\subsection{Experiments}\label{subsec:laval_experiments}

We evaluated both the NEA and RC-EA algorithms on the Laval city graph.  We ran three sets of NEA experiments representing different perspectives: the operator perspective ($\alpha=0.0$), the passenger perspective ($\alpha=1.0$), and a balanced perspective ($\alpha=0.5$).  We only ran RC-EA at $\alpha=1.0$ because as we observed previously in \autoref{sssec:rcea}, RC-EA performs poorly at $\alpha<1.0$.  As in \autoref{sec:mumford_exp}, we perform ten runs of each experiment with ten different random seeds and the same ten different learned policies $\pi_\theta$ as in our other experiments, and we report statistics over these ten runs.  

We kept the same parameters of NEA and RC-EA that we used in \autoref{subsec:baseline}: $IT=400$, $B=10$, $E=10$, and a $300$s transfer penalty.  Although the computer memory requirements of running our algorithms on this very large graph were greater than for the Mumford benchmark datasets, the desktop machine used for our Mumford experiments proved adequate for the Laval experiments as well, and no changes to our model or algorithm were required to reduce memory usage.

\subsubsection{Results}\label{sssec:laval_results}

\begin{figure}
    \centering
    \includegraphics[width=0.6\columnwidth]{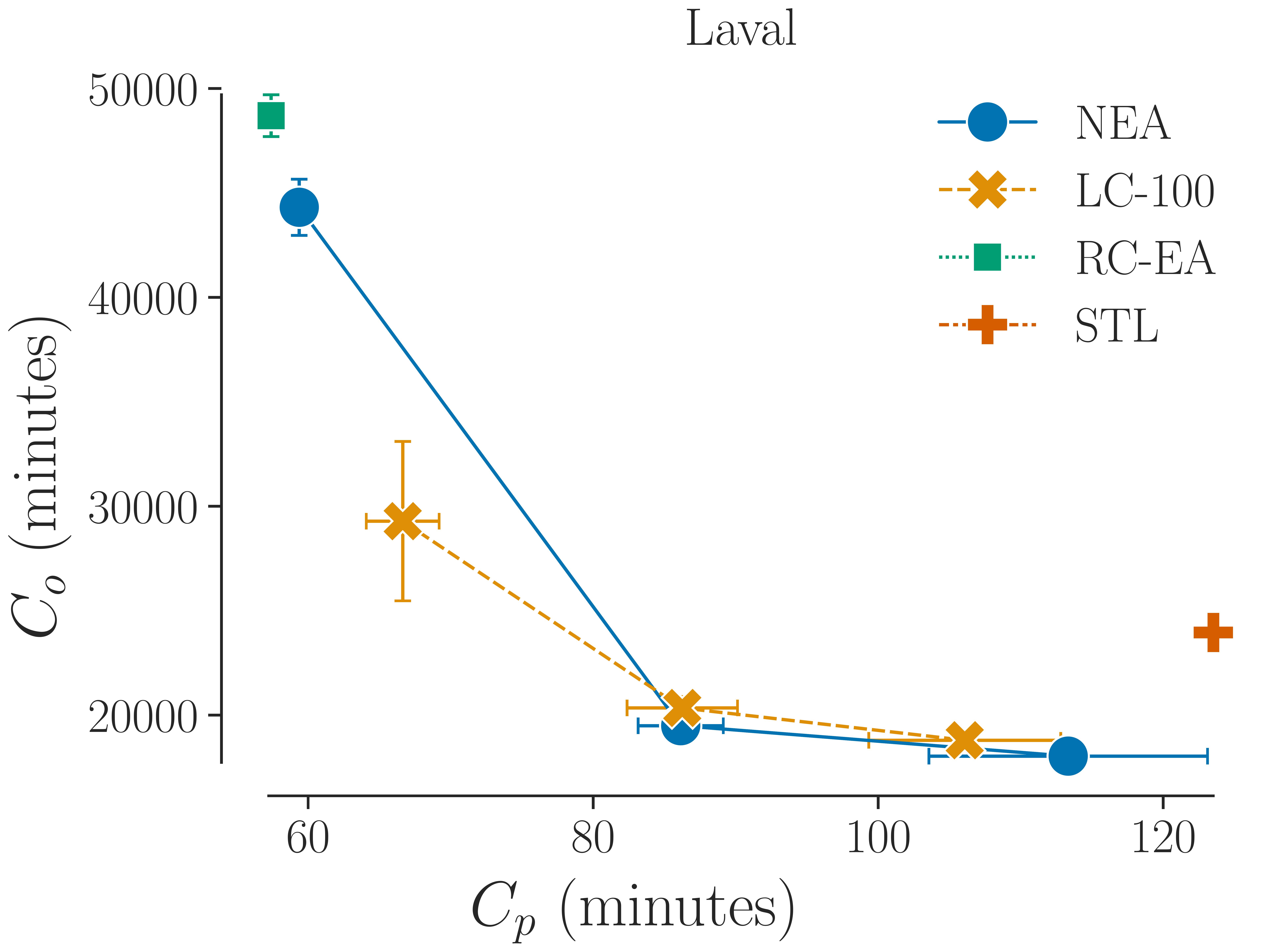}        
    \caption{Average trip time $C_p$ versus total route time $C_o$ achieved by LC-100 and NEA at $\alpha=0.0$ (bottom-right), 0.5 (middle), and 1.0 (top-left), as well as for RC-EA at $\alpha=1.0$ and the STL network (independent of $\alpha$).  Except for STL, each point is a mean value over 10 random seeds, and bars show one standard deviation.}
    \label{fig:laval_pareto}
\end{figure}

The results of the Laval experiments are shown in \autoref{tab:laval_results}.  The results for the initial networks produced by LC-100 are presented as well, to show how much NEA improves on these initial networks.  \autoref{fig:laval_pareto} highlights the trade-offs each method achieves between average trip time $C_p$ and total route time $C_o$.  

\begin{table*}[]
    \centering
    \caption{Performance of LC-100 and NEA's networks at three $\alpha$ values, RC-EA's networks at $\alpha=1.0$, and the STL network.  Bold values are the best for the corresponding $\alpha$, except where STL performs best at that $\alpha$, in which case no values are bold.  $C_p$ is the average passenger trip time. $C_o$ is the total route time.  $d_i$ is the percentage of trips satisfied with $i$ transfers, while $d_{un}$ is the percentage of trips satisfied with 3 or more transfers.  All values are averaged over ten random seeds, with one standard deviation following `$\pm$' where relevant.}    
    \resizebox{1.0\textwidth}{!}{    
    \begin{tabular}{clcccccc}
        \toprule
        $\alpha$ & Method & $C_p \downarrow$ & $C_o \downarrow$ & $d_0 \uparrow$ & $d_1$ & $d_2$ & $d_{un} \downarrow$ \\
        \midrule
          N/A  & STL   &   123.52 &   23954 &        15.75 &       25.86 &      22.87 &         35.51 \\
        \midrule
            & LC-100 & \bf 106.08 $\pm$ 6.74 & 18797 $\pm$ 190 & 15.45 $\pm$ 0.54 & 22.98 $\pm$ 1.34 & 23.71 $\pm$ 1.08 & 37.86 $\pm$ 1.87 \\
            0.0 & NEA & 113.33 $\pm$ 9.78 & \bf 18037 $\pm$ 302 & 15.44 $\pm$ 0.47 & 22.50 $\pm$ 1.72 & 23.23 $\pm$ 1.27 & 38.83 $\pm$ 2.67 \\
        \midrule
            & LC-100 & 86.25 $\pm$ 3.88 & 20341 $\pm$ 549 & 15.04 $\pm$ 0.46 & 26.46 $\pm$ 1.10 & 26.76 $\pm$ 1.18 & \bf 31.74 $\pm$ 2.22 \\
            0.5 & NEA & \bf 86.14 $\pm$ 2.99 & \bf 19494 $\pm$ 389 & 14.89 $\pm$ 0.62 & 26.18 $\pm$ 1.53 & 26.59 $\pm$ 1.69 & 32.34 $\pm$ 2.90 \\
        \midrule
            & LC-100 & 66.64 $\pm$ 2.56 & \bf 29286 $\pm$ 3815 & 17.16 $\pm$ 1.27 & 34.50 $\pm$ 3.49 & 29.03 $\pm$ 1.00 & 19.30 $\pm$ 4.81 \\
            1.0 & NEA & 59.38 $\pm$ 0.52 & 44314 $\pm$ 1344 & 22.08 $\pm$ 0.71 & 38.79 $\pm$ 2.03 & 27.57 $\pm$ 0.75 & 11.56 $\pm$ 2.22 \\
            & RC-EA & \bf 57.40 $\pm$ 0.17 & 48700 $\pm$ 1001 & \bf 23.73 $\pm$ 0.57 & 44.31 $\pm$ 0.71 & 25.20 $\pm$ 0.62 & \bf 6.76 $\pm$ 0.58 \\
        \bottomrule
    \end{tabular}
    }
    \label{tab:laval_results}
\end{table*}

For each value of $\alpha$, NEA's network outperforms the STL network: it achieves 52\% lower passenger cost $C_p$ at $\alpha=1.0$, 25\% lower operating cost $C_o$ at $\alpha=0.0$, and at $\alpha=0.5$ it achieves 30\% lower $C_p$ and 19\% lower $C_o$.  At both $\alpha=0.0$ and $\alpha=0.5$, NEA dominates the existing transit system, achieving both lower $C_p$ and lower $C_o$.

Comparing NEA's results to LC-100, we see that at $\alpha=0.0$, NEA decreases $C_o$ and increases $C_p$ relative to LC-100, and for $\alpha=1.0$, the reverse is true.  In both cases, NEA improves over the initial network on the objective being optimized, at the cost of the other, as we would expect.  In the balanced case ($\alpha=0.5$), we see that NEA decreases $C_o$ by 4\% and $C_p$ by 0.1\% relative to LC-100.  This matches what we observed in \autoref{subsec:baseline}: NEA's improvements over LC-100 tend to be greater in $C_o$ than in $C_p$.  RC-EA, meanwhile, improves on NEA in terms of $C_p$ at $\alpha=1.0$, as it did in our other experiments.  Here, it decreases $C_p$ by two minutes on average, or about 3.3\%.

Looking at the transfer percentages $d_0$, $d_1$, $d_2$, and $d_{un}$, we see that as $\alpha$ increases, the overall number of transfers shrinks, with $d_0$, $d_1$ and $d_2$ growing and $d_{un}$ shrinking.  At $\alpha=1.0$ RC-EA achieves the fewest transfers.  But at both $\alpha=0.5$ and $\alpha=1.0$, NEA also decreases the overall number of transfers relative to the STL network.

The likely aim of Laval's transit agency is to reduce costs while improving service quality, so the `balanced' case, where $\alpha = 0.5$, is the most relevant in reality.  Some simple calculations can give us a rough estimate of savings NEA offers in this case.  The approximate cost of operating a Laval city bus is 200 Canadian dollars (CAD) per hour.  If each route has the same headway (time between bus arrivals) $h$, the number of buses $N_r$ required by route $r$ is $\left \lceil \frac{\tau_r}{h} \right \rceil$.  The total number of buses $N_\mathcal{R}$ required for a transit network $\mathcal{R}$ is then:
\begin{equation}
N_\mathcal{R} = \sum_{r \in \mathcal{R}} N_r \approx \frac{1}{h} \sum_{r \in \mathcal{R}} \tau_r = \frac{C_o}{h}
\end{equation}

The total cost of the system is then 200 CAD$\times N_\mathcal{R}$.  Assuming a headway of 15 minutes on all routes, that gives a per-hour operating cost of 319,387 CAD for the STL network.  By comparison, the networks from NEA with $\alpha = 0.5$ cost 259,920 CAD per hour on average, saving the transit agency 59,500 CAD per hour.

\section{Discussion}\label{sec:conclusion}

The choice of heuristics has a major impact on the effectiveness of a metaheuristic algorithm, and our results show that neural heuristics trained by \acl{DRL} can outperform human-engineered heuristics on the transit network design problem.  Neural heuristics can also be useful in complex real-world transit planning scenarios, where they can improve an existing transit system in multiple dimensions.

Our heuristic-training method has several limitations.  One is that the construction process on which our neural net policies are trained differs from the evolutionary algorithm - an improvement process - in which it is deployed.    We would like to train our neural net policies directly in the context of an improvement process, which may make them better-suited to that process.  

Another limitation is that we only train neural net policies to make one kind of change to a transit network: concatenating multiple shortest paths into a new route and adding it to the network.  In future, we wish to train a more diverse set of policies, perhaps including route-lengthening and -shortening operators that also could be included as heuristics in a metaheuristic algorithm.

It would also be worthwhile to apply our neural heuristics within other metaheuristics than just the evolutionary algorithm developed here.  Multi-objective metaheuristic algorithms could use neural heuristics in tandem with Pareto optimal solution methods such as TOPSIS \cite{papathanasiou2018topsis} to find optimal transit networks.

One remaining question is why the random path-combining heuristic used in the RC-EA experiments outperforms the neural net heuristic for the passenger perspective.  As noted in \autoref{sssec:rcea}, in principle the neural net should be able to learn to imitate a random policy if that gives the best outcome.  We wish to explore in more depth why this does not occur, and whether changes to the policy architecture or learning algorithm might allow the neural heuristic's performance to match or exceed that of the random policy.

Our aim in this paper was to show whether neural heuristics can be used in a lightweight metaheuristic algorithm to improve its performance.  Our results show that they can, and in fact that they can allow a simple metaheuristic algorithm to set new state-of-the-art results on challenging benchmark cities.  This is compelling evidence that they may help transit agencies substantially reduce operating costs while delivering better service to riders in real-world scenarios.

\section*{Acknowledgements}

This work was supported by NSERC under Grant 223292.  We thank the Soci\'et\'e de Transport de Laval for the map data and transportation data they provided to us, and the Agence M\'etropolitaine de Transport of Quebec for the origin-destination dataset that they provided.  This data was essential to the experiments presented in this paper.  And we thank Joshua Katz for acting as a sounding board for ideas and giving feedback on our experimental designs.

\bibliographystyle{tfcad}
\bibliography{references}

{\appendix
\section{Neural Net Architecture}\label{apx:arch}

In this appendix, we give a detailed description of the architecture of the neural net policy that is used in our neural heuristic.  The policy $\pi_\theta$ is a neural net with three components: a \ac{GAT} `backbone', a halting module $\textup{NN}_{halt}$, and an extension module $\textup{NN}_{ext}$.  The halting and extension modules both operate on the outputs from the \ac{GNN}.  All nonlinearities are ReLU functions.  A common embedding dimension of $d_\text{embed} = 64$ is used throughout; unless otherwise specified, each component of the system outputs vectors of this dimension.

\subsection{Input Features}

The policy net operates on three inputs: an $n \times 4$ matrix of node feature vectors $X$, an $n \times n \times 13$ tensor of edge features $E$, and a global state vector $\mathbf{s}$.

A node feature $x_i$ is composed of the $(x,y)$ coordinates of the node and its in-degree and out-degree in the link graph.

An edge feature $\mathbf{e}_{ij}$ is composed of the following elements:
\begin{enumerate}
    \item $D_{ij}$, the demand between the nodes
    \item $s_{ij} = 1$ if $(i,j,\tau_{ij}) \in \mathcal{E}_s$, $0$ otherwise
    \item $\tau_{ij}$ if $(i,j,\tau_{ij}) \in \mathcal{E}_s$, $0$ otherwise   
    \item $c_{ij} = 1$ if $\mathcal{R}$ links $i$ to $j$, $0$ otherwise
    \item $c_{0ij} = 1$ if $j$ can be reached from $i$ over $\mathcal{R}$ with no transfers, $0$ otherwise
    \item $c_{1ij} = 1$ if one transfer is needed to reach $j$ from $i$ over $\mathcal{R}$, $0$ otherwise
    \item $c_{2ij} = 1$ if two transfers are needed to reach $j$ from $i$ over $\mathcal{R}$, $0$ otherwise
    \item $q_{ij} = 1$ if $i = j$, $0$ otherwise
    \item $\tau_{\mathcal{R}ij}$ if $\mathcal{R}$ links $i$ to $j$, $0$ otherwise
    \item $\tau_{rij}$ if $j$ can be reached from $i$ over $\mathcal{R}$ with no transfers, where $r$ is the route that provides the shortest direct trip between $i$ and $j$.  If $i$ and $j$ are not linked with no transfers, has value $0$    
    \item $T_{ij}$, the shortest-path driving time between $i$ and $j$
    \item $\alpha$
    \item $1-\alpha$
\end{enumerate}

The global state vector $\mathbf{s}_t$ at timestep $t$ is composed of:
\begin{enumerate}
    \item $C_p(\mathcal{G}, \mathcal{R}_t \cup \{r_t\})$, the average passenger trip time for the current route set
    \item $C_o(\mathcal{G}, \mathcal{R}_t \cup \{r_t\})$, the total route time for the current route set
    \item $|\mathcal{R}_t|$, the number of routes planned so far
    \item $S - |\mathcal{R}_t|$, the number of routes left to plan
    \item $F_{un}$, the fraction of node pairs with demand $d_{ij} > 0$ that are not connected by $\mathcal{R}_t \cup \{r_t\})$
    \item $\alpha$
    \item $1-\alpha$
\end{enumerate}

Note that $\alpha$ and $1-\alpha$ are included in both $\mathbf{s}$ and $\mathbf{e}_{ij}$.



\subsection{Backbone network}

The backbone is a graph attention net composed of five GATv2\citep{gatv2conv} multi-head graph attention layers separated by ReLU nonlinearities.  4 heads are used in each multi-head attention layer.  The network takes as input the node feature collection $X$ and the edge features $E$, and the final layer outputs an $n \times d_\text{embed}$ matrix of node embeddings $Y$.  These node embeddings are used by the two policy heads to compute action probabilities.

\subsection{Halting module}

$\textup{NN}_{halt}$ is a simple feed-forward \ac{MLP} with 2 hidden layers.  It takes as input $\mathbf{s}_t$, $\tau_{r_t}$, and $\mathbf{y}_i$ and $\mathbf{y}_j$, where $i$ and $j$ are the first and last nodes on $r_t$.  It outputs a scalar $h$, to which we apply the sigmoid function to get the halting policy: 
\begin{align}
\pi(\textup{halt}|s_t) = \sigma(h) &= \frac{1}{1 + e^{-h}} \\ 
\pi(\textup{continue}|s_t) &= 1 - \pi(\textup{halt}|s_t)
\end{align}

\subsection{Extension module}

The extension module $\textup{NN}_{ext}$ computes the policy on odd-numbered timesteps, when extensions are selected for routes.  It consists of two components.  The first is an MLP called $\text{MLP}_{ext1}$, that computes the node-pair scores $o_{aij}$ that are used to calculate the probability of selecting each extension path $a$ to the current route $r_t$.  It takes as input the time between $i$ and $j$ along the path $p$ being considered, $\tau_{pij}$; the node embeddings $\mathbf{y}_i$ and $\mathbf{y}_j$, and the edge feature $\mathbf{e}_{ij}$, and outputs a score $o_{pij}$.  When considering extending an existing route $r_t$ by a path $a$, that act would connect each pair of nodes on $a$, as well as each node already on $r_t$ to each node on $a$.  So the score $o_a$ is just the sum of the node-pair scores for the node pairs that this choice would link together on the route:
\begin{equation}
    o_{(r_t|a)} = \sum_{i \in a, j \in a, i \neq j} o_{aij} + \sum_{i \in r_t, j \in a} o_{(r_t|a)ij}
\end{equation}

If $\alpha = 0$, then the benefit of choosing a path $a$ to start or extend a route depends on its driving time $\tau_a$, and generally on the state $s$.  And if $0 < \alpha < 1$, then the benefit of $a$ depends on these as well as on the edges it adds to $\mathcal{E}_\mathcal{R}$, and so depends on $o_{(r|a)}$.  We account for this with a second \ac{MLP}, called $\text{MLP}_{ext2}$, that takes as input $\mathbf{s}_t, \tau_a,$ and the previous score $o_{(r|a)}$, and outputs a final scalar score $\hat{o}_a$ for each candidate path.  The extension policy is then the softmax of these values:
\begin{equation}
\pi_\theta(a|s) = \frac{e^{\hat{o}_a}}{\sum_{a' \in \mathcal{A}} e^{\hat{o}_{a'}}}
\end{equation}

We thus treat the outputs of $\textup{NN}_{ext}$ as un-normalized log-probabilities of the possible actions during an extension step.

\subsection{Value Network}

The \ac{PPO} training algorithm requires a learnable value function $V(s_t)$ to compute the advantage $A_t = G^H_t - V(s_t)$ during training.  We parameterize $V(s_t)$ as an \ac{MLP} with 2 hidden layers of dimension 36.  The inputs to $V(s_t)$ are the average of the node feature vectors $\frac{\sum_{i}\mathbf{x}_i}{n}$, the total demand $\sum_{i,j} D_{ij}$, the means and standard deviations of the elements of $D$ and $T$, the cost component weight $\alpha$, and the state vector $\mathbf{s}_t$.

We experimented with using an additional neural net head that would compute the value estimate $V(s_t)$ based on the same node embeddings $Y$ as $\textup{NN}_{ext}$ and $\textup{NN}_{halt}$, but found that this offered no benefit over the version described here.

\section{Training Hyper-Parameters}

Training of the neural net policies is performed using the \ac{PPO} \acl{RL} algorithm in conjunction with the well-known Adam optimizer~\citep{kingma2015adam} to update the neural net's parameters based on the losses given by \ac{PPO}. 
 \autoref{tab:learning_params} gives the hyper-parameters used during training.  We augment the training dataset by applying a set of random transformations In each iteration of training.  These include rescaling the node positions by a random factor uniformly sampled in the range $[0.4, 1.6]$, rescaling the demand magnitudes by a random factor uniformly sampled in the range $[0.8, 1.2]$, mirroring node positions about the $y$ axis with probability $0.5$, and revolving the node positions by a random angle in $[0, 2\pi)$ about their geometric center.   

We also randomly sample $\alpha$ separately for each training city in each batch.  Each time alpha is sampled, it has equal probability of being set to $0.0$, set to $1.0$, or sampled uniformly in the range $[0,1]$.  This is to encourage the policy to learn to handle not only intermediate cases, but also the extreme cases where only $C_p$ or only $C_o$ matters.

Although it is typical to use an entropy-maximization term in the loss function when using \ac{PPO}, we found better results by dropping this term, so we do not use it when training out policies.

A set of preliminary experiments with a range of values for the discount rate $\gamma$ led us to set this parameter to $\gamma=0.95$. 

\begin{table}[]
    \centering
    \caption{Training Hyper-Parameters}
    \begin{tabular}{lr}
        \toprule
        Hyper-Parameter & Value \\
        \midrule
        Baseline model learning rate & $5 \times 10^{-4}$ \\
        Baseline model weight decay & 0.01 \\
        Policy learning rate & 0.0016 \\
        Policy weight decay & $8.4 \times 10^{-4}$ \\
        Number of training iterations & 200 \\
        Discount rate $\gamma$ & 0.95 \\
        PPO CLIP threshold $\epsilon$ & 0.2 \\
        Generalized Advantage Estimation $\lambda$ & 0.95 \\
        Batch size & 256 \\
        Horizon & 120 \\
        Num. epochs & 1 \\
        Constraint weight $\beta$ & 5.0 \\
        $S$ & 10 \\
        $MIN$ & 2 \\
        $MAX$ & 12 \\
        Adam $\alpha$ & 0.001 \\
        Adam $\beta_1$ & 0.9 \\
        Adam $\beta_2$ & 0.999 \\
        \bottomrule
    \end{tabular}
    \label{tab:learning_params}
\end{table}
}


\begin{acronym}
  \acro{AV}{autonomous vehicle}
  \acro{TSP}{travelling salesman problem}
  \acro{VRP}{vehicle routing problem}  
  \acro{CVRP}{Capacitated \ac{VRP}}
  \acro{NDP}{transit network design problem}
  \acro{FSP}{Frequency-Setting Problem}
  \acro{DFSP}{Design and Frequency-Setting Problem}
  \acro{SP}{Scheduling Problem}
  \acro{TP}{Timetabling Problem}
  \acro{NDSP}{Network Design and Scheduling Problem}
  \acro{HH}{hyperheuristic}
  \acro{SA}{Simulated Annealing}
  \acro{ACO}{ant colony optimization}
  
  \acro{MoD}{Mobility on Demand}
  \acro{AMoD}{Autonomous Mobility on Demand}
  \acro{IMoDP}{Intermodal Mobility-on-Demand Problem}
  \acro{OD}{Origin-Destination}
  \acro{CSA}{Connection Scan Algorithm}

  \acro{CO}{combinatorial optimization}
  \acro{NN}{neural net}
  \acro{RNN}{recurrent neural net}
  \acro{ML}{Machine Learning}
  \acro{MLP}{Multi-Layer Perceptron}
  \acro{RL}{reinforcement learning}
  \acro{DRL}{deep reinforcement learning}
  \acro{GAT}{graph attention net}  
  \acro{GNN}{graph neural net}
  \acro{MDP}{Markov Decision Process}
  \acro{DQN}{Deep Q-Networks}
  \acro{ACER}{Actor-Critic with Experience Replay}
  \acro{PPO}{Proximal Policy Optimization}
  \acro{ARTM}{Metropolitan Regional Transportation Authority}
  \acro{CDA}{census dissemination area}
\end{acronym}

\end{document}